\documentclass[journal,onecolumn]{IEEEtran}
\ifCLASSINFOpdf
\else
\fi
\usepackage{epsfig}
\usepackage{graphicx}
\usepackage{amsmath}
\usepackage{amssymb}
\usepackage{commath}
\usepackage{bbding}
\usepackage{cite}
\usepackage{cleveref}
\usepackage{booktabs,multirow}

\usepackage{epstopdf}

\begin{document}
% \title{Feature-Independent Projective Image Registration in the Haar Wavelet Domain}
% \title{Projective Image Registration in the Haar Wavelet Domain Without Feature Detection and Matching}
% \title{Feature-Independent Projective Image Registration in the Haar Domain}
%\title{Projective Image Registration in the Haar Domain Without Tracking Features}
% \title{Projective Image Registration in the Haar Wavelet Domain without Corresponding Features}
%\title{Direct Projective Image Registration in the \\Haar Wavelet Domain}
% \title{Projective Image Registration Without Point Correspondences in the Haar Domain}
%--------------------------------------------------------------------------------------------- 
%--------------------------------------------------------------------------------------------- 
%--------------------------------------------------------------------------------------------- 
\title{Image Annotation using Multi-Layer Sparse Coding}

\author{Amara~Tariq\thanks{Amara Tariq is with the Department of Computer Science, Forman Christian College, Pakistan (e-mail:amaratariq@fccollege.edu.pk)} and~Hassan~Foroosh
        % <-this % stops a space
\thanks{Hassan Foroosh is with the Department of Computer Science, University of Central Florida, Orlando,
FL, 32816 USA (e-mail: foroosh@cs.ucf.edu).}% <-this % stops a space
}

\maketitle

\begin{abstract}
Automatic annotation of images with descriptive words is a challenging problem with vast applications in the areas of image search and retrieval. This problem can be viewed as a label-assignment problem by a classifier dealing with a very large set of labels, i.e., the vocabulary set. We propose a novel  annotation method that employs two layers of sparse coding and performs coarse-to-fine labeling. {\it Themes} extracted from the training data are treated as coarse labels. Each {\em theme} is a set of training images that share a common subject in their visual and textual contents. Our system extracts coarse labels for training and test images without requiring any prior knowledge. Vocabulary words are the fine labels to be associated with images. Most of the annotation methods achieve low recall due to the large number of available fine labels, i.e., vocabulary words. These systems also tend to achieve high precision for highly frequent words only while relatively rare words are more important for search and retrieval purposes. Our system not only outperforms various previously proposed annotation systems, but also achieves symmetric response in terms of precision and recall. Our system scores and maintains high precision for words with a wide range of frequencies. Such behavior is achieved by intelligently reducing the number of available fine labels or words for each image based on coarse labels assigned to it.

\end{abstract}



\section{Introduction}
\label{sc:intro}
Automatic annotation of images with accurate textual labels or words is a challenging yet important problem. Such systems have potentially vast applications in the areas of image search and retrieval where search engines have to retrieve appropriate images in response to textual queries of users. The image annotation process can be viewed as a label-assignment problem where a classifier has to choose appropriate labels or words for each image from a large variety of labels, i.e., the set of vocabulary words. Therefore, we use the terms `word' and `label' interchangeably in this paper. 

Performance of image annotation systems is generally measured in terms of mean precision-per-word and mean recall-per-word scores. Precision$(P)$ and recall$(R)$ scores are dependent on $true\;positive(TP)$, $false\;positive(FP)$ and $false\;negative(FN)$ values.

\begin{equation}
P = \frac{TP}{TP + FP},\;\;\;\;\;\;\;
R = \frac{TP}{TP + FN},
\label{eq:rec}
\end{equation}

Let us assume that an annotation system is based on a random classifier. This classifier chooses a set of unique labels $\mathcal{Z}$ such that $z=|\mathcal{Z}|$ for a given image from the set of all available labels, i.e., the vocabulary set $\mathcal{W}$ where $M=|\mathcal{W}|$ ($\mathcal{Z}\subset\mathcal{W}$). The number of all possible subsets of size $z$ is given by
\begin{equation}
Z_{all} = {{M}\choose{z}} = \frac{M!}{z!(M-z)!}
\end{equation}
If $Z_{l}$ denotes the number of all possible subsets of size $z$ containing a certain label/word $w_l$, then 
\begin{equation}
Z_{l} = {{M-1}\choose{z-1}} = \frac{(M-1)!}{(z-1)!(M-z)!}
\end{equation}
Let $p(w_l)$ denote the probability of assigning label/word $w_l$ to the image  while $\hat{p}(w_l)$ is the probability of not assigning this label/word to the image. 
\begin{equation}
p(w_l) = \frac{Z_l}{Z_{all}} = \frac{(M-1)!}{(z-1)!(M-z)!} \times \frac{z!(M-z)!}{M!}
\end{equation}
\begin{equation}
p(w_l) = \frac{z}{M}
\end{equation}
\begin{equation}
\hat{p}(w_l) = 1-p(w_l) = \frac{M-z}{M}
\end{equation}
If $X$ is the fraction of all images that have label/word $w_l$ as their true label, then probabilities of $true\;positive$, $false\;positive$ and $false\;negative$ are  as follows.
\begin{equation}
p(TP) = \frac{zX}{M}
\end{equation}
\begin{equation}
p(FP) = \frac{z(1-X)}{M}
\end{equation}
\begin{equation}
p(FN) = \frac{X(M-z)}{M}
\end{equation}
In general, $z << M$, i.e., only a few labels/words are assigned to each image whereas a large number of words are part of the vocabulary set. Using all these relations in Equation \ref{eq:rec}, we gain the following insight into the precision and recall scores.
\begin{equation}
P \propto X, \;\;\;\;\; \mbox{whereas} \;\;\;\;\; 
R \propto \frac{1}{M}
\label{eq:rec2}
\end{equation}
For a simple annotation system based on a random classifier, 
\begin{itemize}
\item Precision for a word is directly proportional to its frequency, i.e., highly frequent words or labels tend to get better precision scores. 
\item Recall for a word is inversely proportional to the size of the labels' set, i.e., large number of available words or labels causes lower recall scores.
\end{itemize}

The above analysis explains the behavior of various previously proposed annotation system. Most of the previously proposed systems achieve much lower recall score than the precision score. This type of imbalanced response is rooted in the large vocabulary or labels' set available to image annotation systems. Moreover, image annotation systems achieve better precision scores for highly frequent words. Research in the field of text search and retrieval concluded that relatively rare words are more important in search and retrieval scenario than highly frequent words\cite{vsm_text_retrieval}. In terms of information theory, highly frequent words are the {\em ``expected"} events and rare words are the {\em ``surprising"} events. {\em Surprising} events or rare words have more information content than {\em expected} events. 

In this paper, we propose a novel image annotation system that overcomes the above mentioned shortcomings in the response of previously proposed image annotation systems. Our system employs multiple layers of sparse coding and performs coarse-to-fine labeling of images. System identifies distinct {\em themes} present in the training data without requiring any additional prior knowledge and uses these {\em themes} as coarse labels for test images. Each {\em theme} is defined by a set of training images that share a common subject in their visual and textual contents. Vocabulary words are the fine labels which are then assigned to the test image in light of the coarse labels of {\em themes} assigned to it.
 
%% combine
%In this paper, we propose a novel image annotation system that performs coarse-to-fine labeling of images and employs multiple layer of sparse coding. The proposed system is built to overcome the above mentioned shortcomings found in the response of previously proposed systems. The system identifies a set of {\em themes} from the available training data without requiring any prior knowledge or additional input. These {\em themes} are the coarse labels and individual vocabulary words are the fine labels. Each {\em theme} is defined by a set of training images that share a common subject in their visual and textual contents. When the proposed system is given a test image for annotation, it first identifies appropriate {\em themes} for this image at the first layer of sparse coding. Vocabulary set is filtered such that it contains only the words relevant to the {\em theme} of the given test image. The next layer of sparse coding selects the appropriate words or labels for the given test image from a reduced vocabulary set. The reduction in the labels' set aids the proposed system in achieving high recall score. {\em Theme}-based reduction in vocabulary set ensures that the system focuses on the words relevant to the {\em theme}, even when such words are relatively rare in the overall dataset. This characteristic ensures that the system maintains high precision scores for even low frequency or rare words. 

Sparse coding is the process of learning a sparse representation of an input signal in terms of coefficients of a set of basis vectors or predictor variables. Structured sparse coding assumes that there is inherent group structure among predictor variables and employs that structure while modeling input signal. Sparse coding has been widely used in image processing and computer vision communities for problems with small labels' sets such as face recognition and image classification \cite{sg:feature_sel_label, sc:face_recog, sg:class}. Previously proposed sparse coding based annotation systems require explicit specification of coarse labels \cite{sg:multilayer_class, sg:codebook_class2}. Such labels are not explicitly available with all datasets.

Our system employs two layers of sparse coding that use training images as predictor variables. The first layer takes the group structure of the predictor variables into account, in terms of the {\em themes} available in the training data. This layer assigns coarse labels or {\em themes} to test images. The second layer deal with only images and labels of the {\em themes} relevant to the given test image. Our system increases its scope of application by identifying coarse labels without requiring any prior knowledge. 

Sparse coding is the process of learning a sparse representation of an input signal in terms of coefficients of a set of basis signals or predictors. Structured sparse coding indicates that the basis vectors or predictors have inherent group structure which is taken into account while learning sparse representation of the target signal. The idea of sparse coding has been widely used in the field of image processing for feature selection and codebook generation\cite{sp:codebook3,sg:feature_sel_label, sp:codebook_class1} as well as face recognition and image classification\cite{sc:face_recog,sg:class}.  In case of feature selection or codebook generation, each visual feature is treated as a basis vector or a predictor and the goal of the system is to select important visual feature. In many face recognition and image classification systems, training images are treated as basis vectors or predictors. The goal of the system is to find out which class of training images matches most closely with the given test image. Classification systems usually deal with a small number of class labels while image annotation is a large labels' set problem where each item is assigned multiple labels. Some systems incorporate class or coarse labels in the process of sparse learning which are not readily available for image annotation datasets\cite{sg:multilayer_class, sg:codebook_class2}. Our system increases its scope of application by identifying coarse labels without requiring any prior knowledge. 

The proposed system employs two layers of sparse coding. The first layer treats all training images as predictors and the test image as the target signal. It incorporates the group structure among predictors, i.e., the {\em theme}-based groups of training images. It learns a sparse representation of the test images over grouped training images and determines the groups or {\em themes} most relevant to the the test image as their coarse labels. The second layer only processes the training images and the vocabulary words of the relevant {\em themes}. The relevant vocabulary set is smaller than the overall vocabulary set and constitutes the intelligently reduced set of available labels that improves the recall score of the system. For a word $w$ of the relevant vocabulary set, sparse representation of the test image is estimated over the training images of the relevant {\em themes} tagged with word $w$. Quality of the learned representation indicates whether or not the test image should be associated with the word $w$. We define a cosine similarity based measure to assess the quality of the learned sparse representation. 

Mean precision per word ($P$) and mean recall per word ($R$) are used as evaluation measures for image annotation system. In general, one of these measures can be improved at the cost of the other. F-score incorporates the trade-off between precision and recall and is defined as harmonic mean of precision and recall. 
\begin{equation}
F = \frac{2PR}{P+R}
\end{equation}
Various previously published papers regarding image annotation systems report results in terms of F-score. Our system is designed to overcome the tendency of low recall for annotation systems without sacrificing precision. Hence, it outperforms other systems in terms of F-score. Our system also maintains its high precision score for words with wide range of frequencies. Thus our system is practically much more precise for high information content words of low frequency than other image annotation systems. We performed thorough experimentation to prove the qualities of the proposed system.

Note that the words of the vocabulary sets are the target `labels' that an automatic image annotation system needs to assign to images. In this paper, we use the terms `word' and `label' interchangeably to refer to the words of the vocabulary set.

The rest of the paper is structured as follows. We discuss literature regarding image annotation and sparse coding in Section \ref{sc:rel_work}. The problem of image annotation is formally defined in Section \ref{sc:problem}. Section \ref{sc:system} describes the overall architecture of the proposed system. Results of experimental evaluation are presented in Section \ref{sc:eval} followed by the concluding remarks in Section \ref{sc:conclusion}.

\section{Related Work}
\label{sc:rel_work}

\subsection{Image Annotation}
Automatic image annotation is the task of predicting individual words as labels for any given image. A variety of approaches have been proposed to build image annotation systems in the past, ranging from relevance models from the domain of machine translation to the nearest neighbor type algorithms. 
Relevance model from the domain of machine translation was adapted to treat visual features as a language that needs to be translated into words\cite{crm, cmrm, mbrm, sparsekernelN}. Relevance model is a generative modeling scheme. Relevance model and support vector machine, generative and discriminative models respectively, are combined in \cite{hybridmodelJ}. Performance of all of these systems tends to be highly imbalanced in terms of precision and recall, indicating that the systems are precise for only highly frequent words. In contrast, our system is designed to maintain its performance for a wide range of word-frequencies. Annotation systems like \cite{nn:09:tagprop,nn:12:ap, nn:09:vote,nn:12:2PKNN,nn:13:rnn} identify `nearest neighbors' of the test image from which to propagate the labels to the test image. Canonical correlation analysis has been used to improve the performance of such nearest-neighbor type systems\cite{crossmediamodelK}. Systems based on nearest-neighbor type algorithms outperform relevance model based systems at the cost of increased computational complexity. 

Many systems were designed to employ object and action recognition systems from the domain of computer vision to identify objects and actions depicted in images. These systems incorporate names of objects and actions as nouns and verbs while describing images\cite{berg:11:babytalk, berg:12:description,berg:12:midge}. This approach relies on the availability of efficient and reliable object and recognition systems which are limited in number for practical unconstrained settings. 

Meta-data of images\cite{lovethyneighborsA} alongwith multiple image characteristics like color and shape\cite{multiviewhessianF}, have been incorporated in image annotation systems.

The problem of image annotations has often been treated as multi-label problem while multiple labels are simultaneously employed in learning classifiers or dictionaries\cite{multilabeldictionaryC, multilabelcategorizationD, multilabelsparsecodingE}.  Our approach is to explore semantic meaning in relations among multiple labels in terms of {\em themes}. Previously, semantic information has been quantified and incorporated in image annotation systems in different ways \cite{sledG, nn:12:2PKNN,structuredinferenceB}. Various modeling tools including hessian regularization\cite{multiviewhessianF}, multi-scale hypergraph\cite{multiscalhypergraphI}, matrix factorization\cite{nmfknnL} and support vector machines\cite{svmM} have been employed in image annotation systems. An annotation system to deal with weakly labeled images is presented in \cite{semisupervisedP}. For a psychological perspective, the problem of image annotation is basically the task of predicting simple `entry-level' words for images. Ordonez et al. thoroughly explored the problem of translating encyclopedic categories into simple words to be assigned to images\cite{entrylevellabelsS}.

Vast resources of social media websites have been employed to assist the task of automatic image annotation\cite{flckr:11:viscontext,flckrdist:12}. Blei et al. proposed the idea of modeling documents in a collection as mixture of latent variables through latent Dirichlet allocation (LDA)\cite{TM}. Various image annotation systems were based on LDA modeling of image databases\cite{ls:10:mmlda, ls:13:coglda}. In addition to visual contents, context can also aid the process of automatic image annotation. This idea is the basis for some context-sensitive image annotation systems that derive context information from various sources\cite{sceneAIA, tuckerAIA}. These systems employ computationally efficient relevance model and outperform various computationally more expensive systems.

\subsection{Deep Convolutional Neural Networks}
In recent years, deep convolutional neural networks (CNN) trained over ImageNet\cite{imagenet} have gained tremendous popularity for learning image representations for object identification\cite{AlexNet, VGGNet}. Various systems that focus on the task of retrieving or producing sentence-like captions for images, use deep CNN and recurrent neural networks or long short-term memory networks\cite{neuraltalk, showattendtell}. Image annotation systems can still fulfill the needs of image search and retrieval systems by providing tags to be matched against users' queries while avoiding the time and computational complexity to generate sentence-like captions. Kiros et al. used deep representation of images for automatic image annotation\cite{deep_rep_annotation,deeprepresentationsO}. While convolutional neural networks like \cite{AlexNet} and \cite{VGGNet} were designed to identify a single object label for images, deep and recurrent neural networks have been employed for multi-label ranking and classification problem\cite{deeprankingQ, cnnrnnR}. Training a deep neural network requires availability of large training dataset, and is computationally extremely expensive. Our system employs only pre-trained deep CNN for representation learning.

\subsection{Sparse Coding}
Sparse coding is the process of learning a sparse representation of a signal in terms of coefficients of a set of basis signals or predictor variables. Tibshirani proposed LASSO (least absolute shrinkage and selection operator) that includes an $\ell_1$-norm penalty to induce sparsity and interpretability in the learned model \cite{lasso}. Variations of this model have been proposed to incorporate any inherent structure among the predictor variables \cite{sglasso}. Such modeling is beneficial when the group structure carries semantic meaning. 

The idea of sparse modeling has been explored widely in computer vision and image processing communities for feature selection and codebook generation\cite{sp:codebook3, sg:feature_sel_label, sp:codebook_class1,  sp:codebook_image_regions, sp:codebook2,  sg:feature_sel4}. %sp:codebook1,sg:feature_sel1, sg:feature_sel2, sg:feature_sel3,
Most of these systems treat individual visual features as predictor variables. Image classification and face recognition systems that deal with only a small number of class labels, incorporate sparse modeling by treating training images as predictor variables\cite{sc:face_recog,sg:multilayer_class}. One of the datasets used to evaluate the system proposed in \cite{sc:face_recog} contains faces of $38$ individual. Hence, the size of the class labels' set is $38$. Image annotation is a large labels' set problem with at least a few hundred words or labels in almost every popular image annotation dataset. Each item or image in image annotation datasets has multiple labels which is generally not the case with image classification and face recognition type problems. Several image processing systems incorporate some type of classification labels in the process of sparse coding. For example, systems dealing with image classification concurrently with image annotation are limited in their application as they require training data to have class labels in order to induce group structure in predictor variables\cite{sg:multilayer_class}. 

Sparse coding models learn sparse representation of a signal in terms of coefficients of a set of basis signals/ predictors. Tibshirani proposed LASSO (least absolute shrinkage and selection operator) that induces sparsity and interpretability in the learned model through $\ell_1$-norm penalty\cite{lasso}. Variations of this model have been proposed to incorporate any inherent structure among the predictors, carrying semantic meaning \cite{sglasso}. Sparse coding has been previously used for feature selection, codebook generation and face recognition\cite{sp:codebook1, sp:codebook3,sg:feature_sel_label,sc:face_recog,sg:multilayer_class, sp:codebook_class1}. Most of the feature selection and codebook generation methods based on sparse learning treat individual visual features as predictor variable\cite{sp:codebook3,sp:codebook1}. Some image classification methods dealing with small labels' set treat images as predictor variables\cite{sc:face_recog}. Multilayer sparse coding systems have been used to classify and annotate images concurrently, but require class labels\cite{sg:multilayer_class} to induce group structure among predictors. 

The proposed system employs a multi-layer sparse coding framework inspired by \cite{lasso} that %and deals with a large labels' set classification problem, i.e., image annotation where the vocabulary set acts as the labels' set and each image has multiple labels. The system 
treats images as predictor variables and performs coarse-to-fine or {\em theme}-to-word labeling. It extracts coarse labels, i.e., {\em themes} from the training data without requiring any prior knowledge regarding image classes.

\section{Problem Formulation and Notations}
\label{sc:problem}
An image annotation system is given an image $I$ and returns a set of appropriate words ($V^I=\{v_1, v_2, ... , v_B\}$) for that image. Training set $\mathcal{J}$ contains image-description pairs. Vocabulary set $\mathcal{W}$ contains words used in training images' descriptions ($M = |\mathcal{W}|$, $N =  |\mathcal{J}|$). %Vocabulary set $\mathcal{W}$ is the large set of available labels that causes  annotation systems to achieve low recall values. 
Our system splits the training data $\mathcal{J}$ into non-overlapping clusters such that all images of $k^{th}$ cluster $C_k\in \mathcal{C}$ represent a certain {\it theme} in their visual and textual contents where $\mathcal{C}$ denotes the set of all {\it themes} ($K = |\mathcal{C}|$).

The first layer of sparse coding identifies a set $\mathcal{C}^I \subset \mathcal{C}$ of {\it themes} related to the image $I$. The second layer of sparse coding is only fed the subset of training images $\mathcal{J}^I \subset \mathcal{J}$ where $\mathcal{J}^I = \{J | J \in C_k \land C_k \in \mathcal{C}^I\}$. The subset $\mathcal{W}^I \subset \mathcal{W}$ contains the words used in the descriptions of each $J \in \mathcal{J}^I$. %This reduced labels' set or the relevant vocabulary set helps our system achieve high recall while still being highly precise. 
The second layer of the system builds a fix-sized set $V^I=\{v_1, v_2, ... , v_B\}$ of appropriate tags for image $I$ ($V^I \subset \mathcal{W}^I \subset \mathcal{W}$).

\section{System Architecture}
\label{sc:system}
In this section, we explain the details of each stage of the proposed system for automatic image annotation. %The proposed system involves a pre-processing stage in addition to the multi-layer sparse coding framework for annotation prediction.

\subsection{Preprocessing Stage}
\label{subsc:prep}
At the initial stage of the system, image representations are learned and {\em themes} among training items are identified such that the {\em theme}-structure and image representation scheme complement each other.%. Our system design ensures that the {\em theme}-structure among the images and their chosen visual representations complement each other.

\subsubsection{Visual feature extraction}

Hand-crafted visual features based on low-level color and texture characteristics of images have been traditionally used for image classification and object detection. In some cases, high-level labels like detected objects are used as image representations for image annotation\cite{berg:11:babytalk, berg:12:midge}. Recently, multilayer convolutional neural networks have been employed to automatically learn complex high dimensional mappings for input images, optimized to perform tasks like digit recognition and object detection. Such deep convolutional neural networks constitute end-to-end systems that process raw images, alleviating the need to define and extract suitable visual features for the task at hand.

LeCunn et al. first proposed such deep convolutional neural network (CNN) for digit recognition\cite{CNN}. Deep CNN architectures, inspired by the network proposed in \cite{CNN}, have been designed for object recognition\cite{AlexNet, VGGNet}. CNN requires large sets of labeled data for proper training. Availability of ImageNet database provided such training set for object recognition. ImageNet is a very large database of images in which images are classified based on the objects they show\cite{imagenet}. A subset of ImageNet database containing images from $1000$ different classes, has been popularly used to train CNNs for object recognition. Most of these classes represent common objects like `car', `bicycle', etc. %A variety of systems employ CNN pre-trained over this subset of ImageNet to extract visual features from images, and use the extracted visual features for further processing such as automatic caption generation\cite{neuraltalk}.

We employed the CNN proposed in \cite{AlexNet}, trained over the ImageNet subset, for learning image representations. This CNN architecture included several convolutional, pooling and fully-connected layers. Each image is represented  by $4096$ coefficients of the last fully-connected (fc7) layer. Note that such pre-trained model is suited to extract image representations for image annotation as common object names used as ImageNet class labels, constitute a significant portion of the vocabulary used in image annotation datasets like IAPR TC-12 and ESP.  %ImageNet classes used for pre-training CNN also mainly consist of common object names. 
We observed that approximately $60\%$ to $65\%$ of the vocabulary words for IAPR TC-12 and ESP datasets are either part of the ImageNet labels' set or are synonymous with some ImageNet label used for training CNN \cite{AlexNet}. %In the following section, we will explain how such CNN-learned visual features complement the {\em theme} structures of the image annotation datasets.

\subsubsection{Theme structure among training images}
\label{subsubsc:theme}
According to the concepts of information theory, {\em ``surprising"} events, i.e., rare words have more information content than the {\em ``expected"} events or the highly frequent words. This observation is also reinforced in the field of text mining where relatively rare words are deemed more important for search and retrieval than highly frequent words. Rare words are the distinctive properties of the documents they occur in\cite{vsm_text_retrieval}. For representing textual documents in a manner that assigns important and distinctive words higher weights than other words, {\em tfIdf} text representation was proposed. 

\begin{figure}[h]
\begin{center}
\includegraphics[width = 0.18\textwidth]{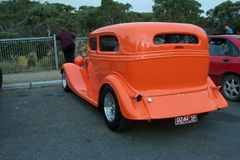}
\includegraphics[width = 0.18\textwidth]{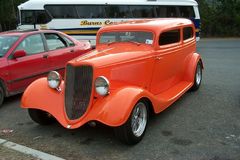}
\includegraphics[width = 0.18\textwidth]{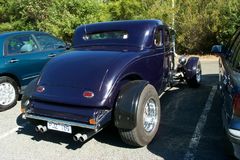}
\includegraphics[width = 0.18\textwidth]{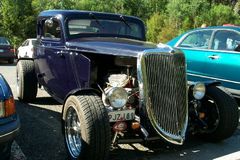}
\includegraphics[width = 0.18\textwidth]{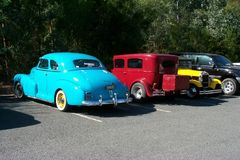}\\
\vspace{0.05cm}
\includegraphics[width = 0.18\textwidth]{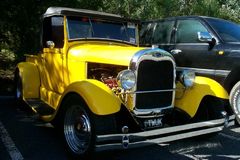}
\includegraphics[width = 0.18\textwidth]{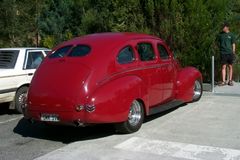}
\includegraphics[width = 0.18\textwidth]{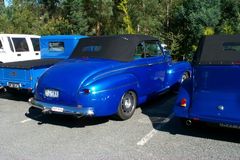}
\includegraphics[width = 0.18\textwidth]{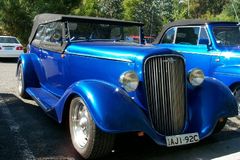}
\includegraphics[width = 0.18\textwidth]{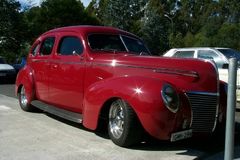}
\end{center}
\vspace{-0.5cm}
\caption{A sample {\it theme} from IAPR TC-12 dataset; `bright colored vintage car' seems to be the distinctive characteristic of this {\it theme}. Ground-truth annotations for these images frequently include words like `vintage', `car', `orange', `red', `yellow', `blue'.}
\label{fg:theme_1}
%\vspace{-0.25cm}
\end{figure}

\begin{figure}[h]
\begin{center}
\includegraphics[width = 0.18\textwidth, height = 0.18\textwidth]{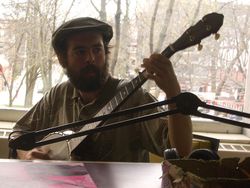}
\includegraphics[width = 0.18\textwidth, height = 0.18\textwidth]{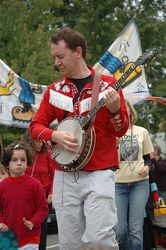}
\includegraphics[width = 0.18\textwidth, height = 0.18\textwidth]{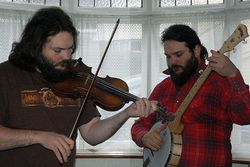}
\includegraphics[width = 0.18\textwidth, height = 0.18\textwidth]{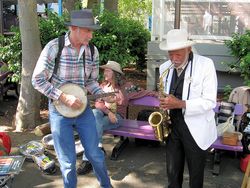}\\
\vspace{0.05cm}
\includegraphics[width = 0.18\textwidth, height = 0.18\textwidth]{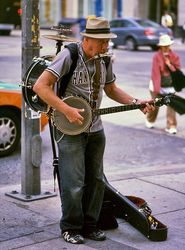}
\includegraphics[width = 0.18\textwidth, height = 0.18\textwidth]{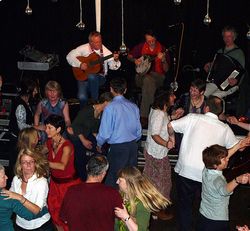}
\includegraphics[width = 0.18\textwidth, height = 0.18\textwidth]{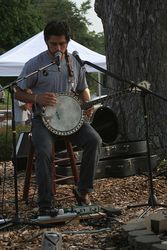}
\includegraphics[width = 0.18\textwidth, height = 0.18\textwidth]{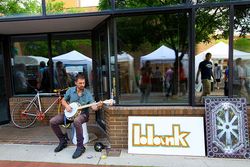}
\end{center}
\vspace{-0.5cm}
\caption{A sample {\it theme} from Flickr30K dataset; ` person playing banjo' seems to be the distinctive characteristic of this {\it theme}. Ground-truth annotations for these images frequently include words like `banjo', `man', `play', `player'.}
\label{fg:theme_2}

\end{figure}

We employ {\em tfIdf} scheme to represent textual descriptions of training images. {\it Tfidf} weight of $i^{th}$ word for $j^{th}$ image description equals $\frac{n_{i}^{j}}{N_i}$ ($n_{i}^{j}$: number of times $i^{th}$ word occurs in $j_{th}$ image description, $N_i$: number of image descriptions that contain $i_{th}$ word.). %We then employ {\em tfIdf} vectors of training image descriptions to define {\em theme} structure among the training data. 
Our system employs cosine similarity based hierarchical clustering to group training images based on {\em tfIdf} representations of their descriptions. Each cluster represents one {\em theme}. This type of clustering ensures that the images sharing the same set of distinctive words, i.e., the words with high {\em tfIdf} scores, are grouped together. Images of one group also tend to share a visual {\it theme}. Figures \ref{fg:theme_1} and \ref{fg:theme_2} show that a clear {\it theme} is apparent in visual contents of images of a group as well. 

We employed similar {\em theme}-structure definition in \cite{tuckerAIA} as well, but this {\em theme} structure also complements image representations in the proposed model. %We argue that the CNN features of images complement the {\em tfIdf}-based {\it theme} structure of the dataset. 
{\it Themes} in training data are determined by the words associated with training images. In conventional image annotation datasets, names of common objects constitute a large portion of these words. A majority of class labels of the ImageNet subset used for training CNN, are the names of everyday objects.  Thus, the {\em theme} structure and the CNN training are both based on words of similar nature. As a result, images of one {\it theme} share substantial similarity in their representations learned by ImageNet-trained CNN.

\subsection{Multi-Layer Sparse Coding}
\label{subsc:multi_lasso}
The proposed system employs two layers of sparse coding to perform coarse-to-fine labeling. The first layer identifies the coarse labels in terms of relevant {\em themes} and the second layer predicts fine labels, i.e, word annotations for images.Both layers treat training images as predictors and the test image as the target signal. The first layer involves group structure among predictor variables or the training images and is responsible for coarse labeling of the test image with its relevant {\em themes}. The second layer produces the final output of the system in terms of fine labels or the words associated with the test image, in light of its relevant {\em themes}.

\subsubsection{Group sparse coding for theme identification}
\label{subsubsc:sglasso}
Tibshirani proposed `least absolute shrinkage and selection operator' or `lasso' for estimation in linear models. Lasso minimizes some coefficients and sets others to exactly $0$. It produces more interpretable models due to such behavior. Lasso has been shown to present favorable characteristics of both subset selection and ridge regression\cite{lasso}.

Many variations of lasso model were proposed to incorporate problem-specific assumptions to improve accuracy. Group lasso assumes that features/predictor variables of a given problem are grouped, and problem has sparse effects both between groups and within groups. Simon et al. proposed a regularized linear regression model with $\ell_1$ and $\ell_2$ norms based penalties to incorporate group structure of predictor variables in the estimation process\cite{sglasso}. The first layer of sparse coding in the proposed system employs similar modeling scheme.

The training images are predictor variables and the test image is the target or the response. {\em Themes} are defined as groups of training images. Hence, the {\em theme} structure of training data defines the group structure of predictor variables required for application of group lasso model.  The goal of this layer is to identify a set of appropriate {\it themes} for the test image. These {\em themes} are the coarse labels assigned to images. Since {\it themes} carry rich semantic and contextual meaning, relevant {\it themes} define a semantic background for the test image. 

Each column of data matrix ($A$) contains CNN features of one training image. Images of the same {\it theme} are placed in adjacent columns.  Test image $I$, in the form of CNN features, is the target variable. The cost is described as
\begin{equation}
\min_{\mathbf{w}}( ||A{\bf w} - I||_{2}^{2} + \lambda_{1}||{\bf w}||_{1} + \lambda_{2}\sum_{k=1}^{K}(\phi_{k}||{\mathbf{w}_{k}}||_2)))
\label{eq:layer1}
\end{equation}
The cost is regularized by two penalties. A penalty based on $\ell_1$-norm of the coefficients vector ${\bf w}$ induces sparsity in the learned model. Sparsity at group-level is induced by $\ell_2$-norm of the groups of coefficients belonging to each {\it theme}. This penalty ensures that as few groups or {\it themes} are assigned non-zero coefficients as possible. Weights assigned to the two penalty terms, i.e., $\lambda_1$ and $\lambda_2$, indicate the emphasis put on each penalty term by the framework. In our approach, equal emphasis is put on all groups or {\it themes}, i.e., $\forall k, \phi_k = 1$

\begin{table*}[t]
\begin{center}
\begin{tabular}{|c||c|}\hline
 Test Image  & Sample images from  {\it themes} related to the test image $I$\\\cline{2-2}
($I$)            &  \hspace{-0.025cm}     {\em theme}-$1$  \hspace{1.9cm}\vrule width 2.25pt\hspace{2cm} {\em theme}-$2$                \\\hline
\includegraphics[width = 0.12\textwidth, height = 0.12\textwidth, trim=0 0 0 -25]{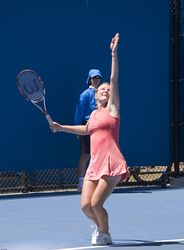} &
\includegraphics[width = 0.12\textwidth, height = 0.12\textwidth, trim=0 0 0 -10]{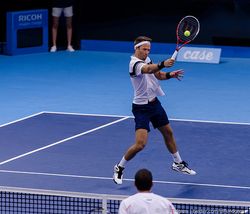} 
\includegraphics[width = 0.12\textwidth, height = 0.12\textwidth, trim=0 0 0 -25]{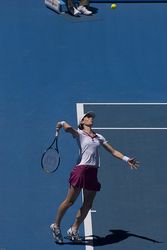} 
\includegraphics[width = 0.12\textwidth, height = 0.12\textwidth, trim=0 0 0 -25]{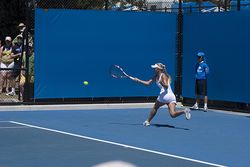} \hspace{0.05cm}\vrule width 2pt\hspace{0.05cm}
\includegraphics[width = 0.12\textwidth, height = 0.12\textwidth, trim=0 0 0 -25]{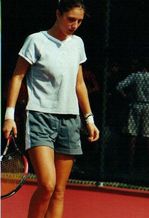} 
\includegraphics[width = 0.12\textwidth, height = 0.12\textwidth, trim=0 0 0 -25]{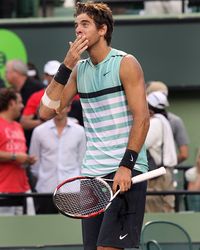} 
\includegraphics[width = 0.12\textwidth, height = 0.12\textwidth, trim=0 0 0 -25]{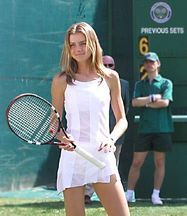}
%\hspace{0.05cm}\vrule width 2pt\hspace{0.05cm}
%\includegraphics[width = 0.08\textwidth, height = 0.08\textwidth, trim=0 0 0 -25]{Figures/multi_themes/95/3/1.jpg} 
%\includegraphics[width = 0.08\textwidth, height = 0.08\textwidth, trim=0 0 0 -25]{Figures/multi_themes/95/3/2.jpg} 
%\includegraphics[width = 0.08\textwidth, height = 0.08\textwidth, trim=0 0 0 -25]{Figures/multi_themes/95/3/3.jpg}
\\\hline

\includegraphics[width = 0.12\textwidth, height = 0.12\textwidth, trim=0 0 0 -25]{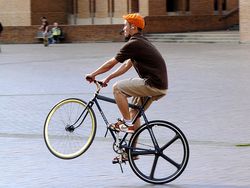} &
\includegraphics[width = 0.12\textwidth, height = 0.12\textwidth,trim=0 0 0 -25]{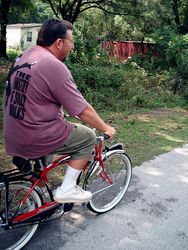} 
\includegraphics[width = 0.12\textwidth, height = 0.12\textwidth, trim=0 0 0 -25]{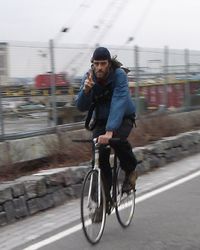} 
\includegraphics[width = 0.12\textwidth, height = 0.12\textwidth, trim=0 0 0 -25]{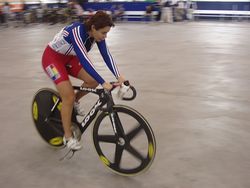} 
\hspace{0.05cm}\vrule width 2pt\hspace{0.05cm}
\includegraphics[width = 0.12\textwidth, height = 0.12\textwidth,trim=0 0 0 -25]{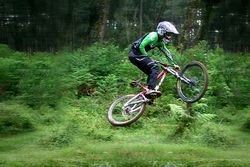} 
\includegraphics[width = 0.12\textwidth, height = 0.12\textwidth,trim=0 0 0 -25]{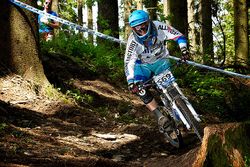} 
\includegraphics[width = 0.12\textwidth, height = 0.12\textwidth,trim=0 0 0 -25]{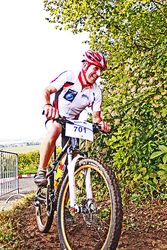}
%\hspace{0.05cm}\vrule width 2pt\hspace{0.05cm}
%\includegraphics[width = 0.08\textwidth, height = 0.07\textwidth, trim=0 0 0 -25]{Figures/multi_themes/902/3/1.jpg} 
%\includegraphics[width = 0.08\textwidth, height = 0.07\textwidth, trim=0 0 0 -25]{Figures/multi_themes/902/3/2.jpg} 
%\includegraphics[width = 0.08\textwidth, height = 0.07\textwidth, trim=0 0 0 -25]{Figures/multi_themes/902/3/3.jpg}
\\\hline

\includegraphics[width = 0.12\textwidth, height = 0.12\textwidth, trim=0 0 0 -25]{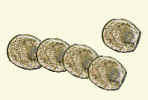} & 
\includegraphics[width = 0.12\textwidth, height = 0.12\textwidth, trim=0 0 0 -25]{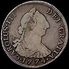} 
\includegraphics[width = 0.12\textwidth, height = 0.12\textwidth, trim=0 0 0 -25]{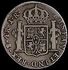} 
\includegraphics[width = 0.12\textwidth, height = 0.12\textwidth, trim=0 0 0 -25]{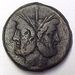} \hspace{0.05cm}\vrule width 2pt\hspace{0.05cm}
\includegraphics[width = 0.12\textwidth, height = 0.12\textwidth, trim=0 0 0 -25]{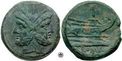} 
\includegraphics[width = 0.12\textwidth, height = 0.12\textwidth, trim=0 0 0 -25]{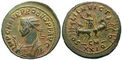} 
\includegraphics[width = 0.12\textwidth, height = 0.12\textwidth, trim=0 0 0 -25]{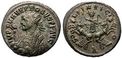}
%\hspace{0.05cm}\vrule width 2pt\hspace{0.05cm}
%\includegraphics[width = 0.08\textwidth, height = 0.07\textwidth, trim=0 0 0 -25]{Figures/multi_themes/15251/3/1.jpg} 
%\includegraphics[width = 0.08\textwidth, height = 0.07\textwidth, trim=0 0 0 -25]{Figures/multi_themes/15251/3/2.jpg} 
%\includegraphics[width = 0.08\textwidth, height = 0.07\textwidth, trim=0 0 0 -25]{Figures/multi_themes/15251/3/3.jpg}
\\\hline

\includegraphics[width = 0.12\textwidth, height = 0.12\textwidth, trim=0 0 0 -25]{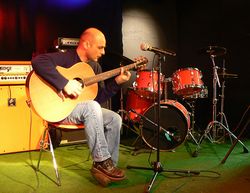} & 
\includegraphics[width = 0.12\textwidth, height = 0.12\textwidth, trim=0 0 0 -25]{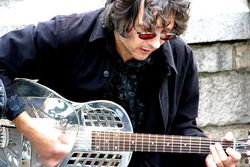} 
\includegraphics[width = 0.12\textwidth, height = 0.12\textwidth,trim=0 0 0 -25]{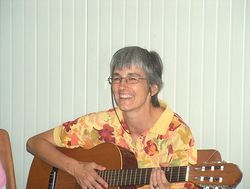} 
\includegraphics[width = 0.12\textwidth, height = 0.12\textwidth, trim=0 0 0 -25]{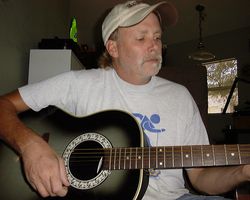} 
\hspace{0.05cm}\vrule width 2pt\hspace{0.05cm}
\includegraphics[width = 0.12\textwidth, height = 0.12\textwidth, trim=0 0 0 -25]{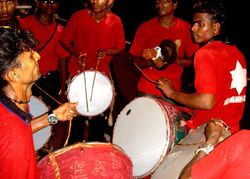} 
\includegraphics[width = 0.12\textwidth, height = 0.12\textwidth, trim=0 0 0 -25]{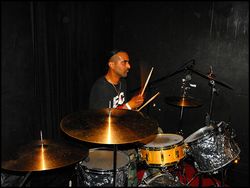}
\includegraphics[width = 0.12\textwidth, height = 0.12\textwidth, trim=0 0 0 -25]{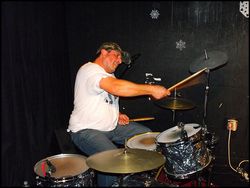}
\\\hline
\end{tabular}
\end{center}
%\vspace{-0.25cm}
\caption{\label{tb:theme_assoc} Each row contains one test image $I$ and sample images from two {\em themes} related to image $I$.}
%\vspace{-0.25cm}
\end{table*}

{\it Themes} corresponding to the coefficient groups with non-zero values are deemed relevant to the test image $I$. Training images of these {\it themes} assigned non-zero coefficient, are put in the subset $\mathcal{J}^{I}$. Words describing these images form the subset $\mathcal{W}^{I}$. In general, $\mathcal{J}^I \subset \mathcal{J}$ and  $\mathcal{W}^I \subset \mathcal{W}$. $\mathcal{W}^I$ is the intelligently reduced labels set, specific to the {\it themes} of test image, that mitigates the adverse effects of the large size of labels set and improves the recall value (Equation \ref{eq:rec2}). To prevent precision values from dropping, it is necessary that the subset $\mathcal{W}^I$ contains all the labels or words that can be potentially related to the image $I$. Association of the image $I$ with multiple {\it themes} avoids unnecessary limiting of the words' or the labels' set. The test image may show characteristics of multiple {\it themes} and should be processed in the light of all these {\it themes}. Each row of Table \ref{tb:theme_assoc} shows one test image and sample images from the {\em themes} related to that test image. For example, test image of the last row of Table \ref{tb:theme_assoc} shows a guitarist and a drum-set. One relevant {\em theme} to this image contains images of guitarists and the other contains images of drum-sets.

\subsubsection{Regularized regression modeling for word prediction}
\label{subsubsc:lasso}
We explained in the preceding section how {\em themes} or coarse labels are assigned to the test image. These coarse labels or {\em themes} define semantic background or context of the test image. The goal of the next layer of sparse coding is to assign fine labels to images in light of their semantic context defined by their relevant {\em themes}. Every word in the vocabulary set defines one fine label. This layer employs a sparse coding framework with one penalty term defined as $\ell_1$-norm of the estimated coefficients. This lasso-inspired modeling scheme is used to learn a model corresponding to every word in the vocabulary for every test image $I$. The implementation details are described as follows.

Test image $I$ is treated as the target signal. Predictor variables are all the training images from set $\mathcal{J}^I$, with word $v$ in their descriptions. These predictors form the set $\mathcal{J}^{I_v} \subset \mathcal{J}^I$ which is representative of $v \in \mathcal{W}^{I}$. Columns of data matrix $A^{I_v}$ correspond to images of $\mathcal{J}^{I_v}$. The cost for word $v$ with image $I$ as target is defined as
\begin{equation}
\min_{\mathbf{w_v}}( ||A^{I_v}{\bf w_v} - I||_{2}^{2} + \rho||{\bf w_v}||_{1})
\label{se:layer2}
\end{equation}
This objective function is a linear regression model with $\ell_1$-norm of coefficients vector $\mathbf{w_v}$ as penalty that induces sparsity in the learned model. 

The model described above tries to encode target signal, i.e., the test image $I$ as a linear combination of the visual representatives of the word $v\in\mathcal{W}^I$. The quality of the estimated encoding of image $I$ hints at the similarity between test image $I$ and visual representation of word $v$. Let $O = A^{I_v}{\bf w_v}$ denotes the reconstruction of image $I$ by the estimated model. We define cosine similarity between $O$ and $I$ as a measure of quality of the estimated model, i.e., the similarity between the test image $I$ and visual representation of word $v$\footnote{Superscript \hspace{1pt}$\hat{}$\hspace{1pt} denotes the normalized vector.}.
\begin{equation}
sim^{I_v} = \hat{O} \cdot \hat{I} =  \frac{(A^{I_v}{\bf w_v})^{T}  {I}} {||A^{I_v}{\bf w_v}|| ||I||}
\end{equation}

A higher value of $sim^{I_v}$ indicates that the representative images of the word $v$ can reconstruct the test image $I$ with low error. It implies that the image $I$ is visually similar to the training images that are tagged with the word $v$. Therefore, image $I$ should also be annotated with the word $v$. One such model is learned for every word $v \in \mathcal{W}^I$. If $V^I$ is the set of $B$ annotations for image $I$, then $V^I$ contains the top $B$ words with respect to their $sim^{I_v}$ scores\footnote{We employed the SLEP software package http://www.yelab.net/software/SLEP/ for the training of the sparse coding models at both layers of the system.}.

%If $B$ words are to be selected for image $I$, top $B$ words with respect to their $sim^{I_v}$ scores are picked\footnote{We employed the SLEP software package http://www.yelab.net/software/SLEP/ for sparse coding training.}.%the training of the sparse coding models at both layers of the system.}. 

\section{Evaluation}
\label{sc:eval}
The proposed method is designed to deal with traditional settings of automatic image annotation problem which involve very large vocabulary sets and lack of availability of any additional labels. Therefore, we used the same datasets and evaluation measures as used by the previously proposed images annotation systems designed to work under these condition such as \cite{mbrm,nn:09:tagprop,nn:12:2PKNN,sceneAIA,deep_rep_annotation,random_forest}.

We employed three popular image annotation benchmark datasets, i.e., the IAPR TC-12\footnote{http://www.imageclef.org/photodata}, the ESP game\footnote{http://hunch.net/ jl/}, and the Flickr30K\footnote{http://shannon.cs.illinois.edu/DenotationGraph/} \cite{flickr30K}. IAPR TC-12 contains $19846$ images, taken by tourists and described in a few sentences. ESP consists of $21844$ images tagged by ESP game. Flickr30K has about $31000$ images, downloaded from Flickr\footnote{www.flickr.com}, showing people engaged in everyday activities. Five captions were collected for each image through crowd-sourcing. 

We used TreeTagger\footnote{3http://www.cis.uni-muenchen.de/ schmid/tools/TreeTagger/} to tokenize, lemmatize and part-of-speech tag IAPR TC-12 and Flickr30K image captions. Frequently occurring nouns, verbs and adjectives are retained (every word occurring in more than one captions for any Flickr30K image is taken as its true label). Sizes of the vocabulary sets for IAPR TC-12, ESP and Flickr30K are $291$, $268$  and $316$, respectively. Random subsets of $10\%$ of the data are used as test sets for IAPR TC-12 and ESP dataset. $1000$ randomly selected items are used to test Flickr30K. The same vocabulary and test sets have been previously used for evaluation for these datasets. 

The IAPR TC-12 dataset contains $19846$ images taken by tourists and carefully described in a few sentences. Each image description is processed through TreeTagger\footnote{3http://www.cis.uni-muenchen.de/ schmid/tools/TreeTagger/} for tokenization, lemmatization and part-of-speech tagging. Most frequently occurring nouns, verbs and adjectives are selected to form a vocabulary set $\mathcal{W}$ of size $291$.  A random subset containing $10\%$ of the dataset is used for testing purposes. The same split of dataset into training and test set as well as the same vocabulary set has been used by various previously proposed methods for evaluation. 

The ESP game dataset contains images tagged with words by the players of the ESP game. A smaller subset of this dataset containing $21844$ items has been previously used as an image annotation benchmark dataset. We used the same subset for fair comparison against previously proposed methods. This subset has a vocabulary set $\mathcal{W}$ consisting of $268$ unique words. Testing has been performed over a randomly picked subset consisting of $10\%$ items of the dataset. Various previously proposed methods have used the same split of the dataset into training and test sets along with the same vocabulary set.

The Flickr30K\footnote{http://shannon.cs.illinois.edu/DenotationGraph/} \cite{flickr30K} dataset contains approximately $31,000$ images collected from Flickr\footnote{www.flickr.com} that depict people engaged in everyday activities. Each image is associated with $5$ captions collected through crowd-sourcing.  This dataset has been previously used to test description retrieval systems and the systems generating sentence-like captions\cite{neuraltalk, showtell,deep_caption_mrnn}. We annotated this dataset with individual words through the proposed method, as well as some previously proposed annotation methods, to evaluate how different methods adapt to datasets of different nature. Testing is performed over $1000$ randomly picked images from the dataset. We used TreeTagger to tokenize, lemmatize, part-of-speech tag captions. Since every image is associated with $5$ captions, every word that is present in more than one caption is taken as a valid annotation for the image. The vocabulary set $\mathcal{W}$ consists of $316$ frequently occurring nouns, verbs and adjectives. 

Training sets of all three datasets are divided into {\em themes} or clusters through the process described in Section \ref{subsubsc:theme}. Clusters with too few members are dropped such that the remaining clusters contain approximately $90\%$ of the dataset. This process results in $771$, $1346$ and $1102$ {\it themes} for IAPR TC-12, ESP game and Flickr30K datasets, respectively.

We used the same evaluation measures as employed by comparative annotation systems proposed in the past, i.e., mean precision per word ($\mathrm{P}$), mean recall per word ($\mathrm{R}$), mean F-measure ($F$)\footnote{Mean F-measure has been defined as the harmonic mean of mean precision and mean recall, i.e., $F =\frac{2\mathrm{P}\mathrm{R}}{\mathrm{P}+\mathrm{R}}$\cite{nn:12:2PKNN}. We use the same definition. } and the number of words with positive recall ($N^+$). 

\subsection{Experimental Results}
\label{subsc:results}

Tables \ref{tb:IAPR}, \ref{tb:ESP} and \ref{tb:Flickr} show performance of various image annotation systems over IAPR TC-12, ESP game and Flickr30K datasets, respectively. Rows of Tables \ref{tb:IAPR} and \ref{tb:ESP} are grouped in terms of the approach of the methods. The first group of rows contain methods based on relevance models. % from the domain of machine translation. 
The second group contains methods using nearest neighbor type algorithms. The third group contains systems using a variety of approaches from greedy label transfer\cite{baseline:08} to random forests \cite{random_forest} and deep neural network based representations \cite{deep_rep_annotation} of images. The fourth group shows performance of some previously proposed systems with deep CNN based image representations. The last row shows the proposed method that is built on deep CNN based representations and multi-layer sparse modeling. 

Flickr30K dataset was introduced relatively recently as compared to IAPR TC-12 and ESP game datasets.  Therefore, various image annotation papers have not reported results over this dataset. We evaluated various previously proposed systems along with our own method, over this dataset.

\begin{table}[t]
\begin{center}
\setlength{\tabcolsep}{4pt}
\begin{tabular}{|c|c|c|c|c|c|}
\hline  & Mean Precision & Mean Recall & Mean F-score   &  $N^+$\\\hline
CRM\cite{crm} & $21$ & $15$ & $18$ & $214$ \\
MBRM\cite{mbrm} & $21$ & $14$ & $17$ & $186$ \\
BS-CRM\cite{bs-crm} & $22$ & $24$ & $23$ & $--$ \\
SKL-CRM\cite{sparsekernelN} & $32$ & $51$ & $39$ & $274$\\
SVM-DMBRM\cite{hybridmodelJ} & $56$ & $29$ & $38$ & $283$\\
Scene-AIA\cite{sceneAIA} & $56$ & $25$ & $35$ & $230$ \\\hline
TagProp-ML\cite{nn:09:tagprop} & $48$ & $25$ & $33$ & $227$ \\
TagProp\cite{nn:09:tagprop} & $46$ & $35$ & $40$ & $266$ \\
FastTag\cite{nn:13:fasttag} & $47$ & $26$ & $34$ & $280$\\
2PKNN-ML\cite{nn:12:2PKNN} & $54$ & $37$ & $44$ & $278$ \\
KCCA+2PKNN\cite{crossmediamodelK} & $59$ & $30$ & $40$ & $259$\\
CCA-KNN\cite{deeprepresentationsO} & $45$ & $38$ & $41$ & $278$\\\hline
JEC\cite{baseline:08} & $25$ & $16$ & $20$ & $196$ \\
Lasso\cite{baseline:08} & $26$ & $16$ & $20$ & $199$ \\
HGDM \cite{hgdm:13} & $29$ & $18$ & $22$ & $--$ \\
AP\cite{nn:12:ap} & $28$ & $26$ & $27$ & $--$ \\
HHD\cite{multiscalhypergraphI} & $32$ & $44$ & $36$ & $280$\\
KSVM-VT\cite{svmM} & $47$ & $29$ & $36$ & $268$\\
Deep rep.\cite{deep_rep_annotation} & $42$ & $29$ & $34$ &$252$\\
Random Forests\cite{random_forest} & $45$ & $31$ & $37$ & $253$\\\hline
Scene-AIA(fc7) & $63$ & $27$ & $38$ & $259$ \\
TagPorp(fc7) & $32$ & $40$ & $36$ & $264$ \\\hline
{\bf MultiSC-AIA(fc7)} & $41$ & $42$ & $42$ & $250$ \\
\hline
\end{tabular}
\end{center}
\vspace{-0.25cm}
\caption{\label{tb:IAPR} Performance evaluation for IAPR-TC-12 dataset }
\vspace{-0.25cm}
\end{table}
\begin{table}[h]
\begin{center}
\setlength{\tabcolsep}{4pt}
\begin{tabular}{|c|c|c|c|c|c|}
\hline  & Mean Precision & Mean Recall & Mean F-score   &  $N^+$\\\hline
CRM\cite{crm} & $29$ & $19$ &  $18$ & $227$\\% & $$ & $$ & $$ & $$ \\%
MBRM\cite{mbrm} & $21$ & $17$ & $19$ & $218$ \\
SKL-CRM\cite{sparsekernelN} & $26$ & $41$ & $32$ & $248$\\
SVM-DMBRM\cite{hybridmodelJ} & $55$ & $25$ & $34$ & $259$\\
Scene-AIA\cite{sceneAIA} &  $60$ & $20$ & $30$ & $234$ \\\hline
TagProp-ML\cite{nn:09:tagprop} &  $49$ & $20$ & $28$ & $213$ \\
TagProp\cite{nn:09:tagprop} &  $39$ & $27$ & $32$ & $239$ \\
FastTag\cite{nn:13:fasttag} &  $46$ & $22$ & $30$ & $247$ \\
2PKNN-ML\cite{nn:12:2PKNN} &  $53$ & $27$ & $36$ & $253$ \\
NMF-KNN\cite{nmfknnL} & $33$ & $26$ & $29$ & $238$ \\
CCA-KNN\cite{deeprepresentationsO} & $46$ & $36$ & $41$ & $260$\\\hline
JEC\cite{baseline:08} & $23$ & $19$ & $21$ & $227$ \\ 
Lasso\cite{baseline:08} & $22$ & $18$ & $21$ & $225$ \\
AP\cite{nn:12:ap} & $24$ & $24$ & $24$ & $--$ \\
HHD\cite{multiscalhypergraphI} & $35$ & $36$ & $34$ & $257$\\
KSVM-VT\cite{svmM} & $33$ & $32$ & $33$ & $259$\\
Deep rep.\cite{deep_rep_annotation} & $38$ & $22$ & $28$ &$228$\\
Random Forests\cite{random_forest} & $45$ & $24$ & $31$ & $239$\\\hline
Scene-AIA(fc7) & $61$ & $21$ & $31$ & $245$ \\
TagPorp(fc7) & $29$ & $37$ & $33$ & $245$ \\\hline
{\bf MultiSC-AIA(fc7)} & $39$ & $38$ & $39$ & $237$ \\
\hline
\end{tabular}
\end{center}
\vspace{-0.5cm}
\caption{\label{tb:ESP} Performance evaluation for ESP dataset }
\end{table}
\begin{table}[h]
\begin{center}
\setlength{\tabcolsep}{4pt}
\begin{tabular}{|c|c|c|c|c|c|}
\hline  & Mean Precision & Mean Recall & Mean F-score   &  $N^+$\\\hline
MBRM & $16$ & $23$ & $19$ & $264$ \\
Scene-AIA(fc7) & $35$ & $18$ & $24$ & $179$ \\
TagPorp(fc7) & $23$ & $28$ & $25$ & $243$ \\\hline
{\bf MultiSC-AIA(fc7)} & $26$ & $27$ & $27$ & $243$ \\
\hline
\end{tabular}
\end{center}
\vspace{-0.25cm}
\caption{\label{tb:Flickr} Performance evaluation for Flickr30K dataset }
\vspace{-0.25cm}
\end{table}

\subsection{Observations}
\label{subsc:obs}
Note that the precision can be increased for many systems at the cost of decrease in recall and vice versa. F-measure is a unified evaluation measure that incorporates the trade-off between precision and recall. As shown in Tables \ref{tb:IAPR}, \ref{tb:ESP} and \ref{tb:Flickr}, our method outperforms almost all of the previously proposed approaches in terms of mean F-score for all datasets, while maintaining both high precision and high recall in a balanced fashion. It indicates that our system maintains its designed characteristics for a wide variety of datasets. 

For IAPR TC-12 dataset, only 2PKNN-ML\cite{nn:12:2PKNN} performs slightly better than our system while being highly imbalanced in terms of precision and recall. The recall performance of our system is still better than corresponding value for 2PKNN-ML. Similar trend is noted for CCA-KNN\cite{deeprepresentationsO} for ESP dataset. The performance of Sprase kernel CRM\cite{sparsekernelN} seems to be imbalanced in the opposite direction, i.e., high recall and low precision. NMF-KNN\cite{nmfknnL} was designed to tackle the problem of imbalance in label frequency, i.e., frequencies of different words. Still, its performance is not balanced in terms of precision and recall. 

%Our method also outperforms all other methods in terms of mean F-measure for ESP game and Flickr30K dataset. It performs comparable to the best performing previously proposed approach (2PKNN-ML) for the IAPR TC-12 dataset, while outperforming all other methods in terms of mean F-measure. The precision and recall values achieved by our system, are very close to each other for all three datasets, indicating that our method maintains its designed characteristics for datasets of different nature. 

A variety of visual features were evaluated in the past by image annotation systems, ranging from grid-based visual features\cite{mbrm} to normalized-cuts based {\it blobs}\cite{cmrm}. Our method includes a deep convolutional neural network at the initial processing stage to automatically learn visual representations for images. Such image representation has greatly improved the performance of image classification and character recognition systems\cite{AlexNet, VGGNet, CNN}. Our approach reaps the rewards of effective image representation learning by using deep convolutional neural networks. 

We tested the previously proposed methods such as Scene-AIA and TagProp with the same image representation (denoted by `fc7') as used by our model.  It is obvious from Tables \ref{tb:IAPR}, \ref{tb:ESP} and \ref{tb:Flickr} that our classifier outperforms previous methods even when they use the same image representation, proving the effectiveness of our sparse coding framework. TagProp is based on a nearest neighbor algorithm and was originally used with images represented by a combination of local and holistic features in \cite{nn:09:tagprop}. For `fc7' features, this system achieves higher recall and comparatively low precision as compared to its application over visual features described in \cite{nn:09:tagprop} for IAPR TC-12 and ESP game datasets. For both types of image representations, precision and recall values are vastly different from each other and the F-measure is less than the value achieved by our classifier.

Table \ref{tb:recall} shows samples of the words assigned high and low recall values for all three datasets. It is evident that, in addition to frequently occurring words, words related to distinct visual {\it themes} or scenes (such as {\em tennis match},  {\em bicycle race}, {\em soccer match}) achieve better recall. This property can be attributed to the {\it theme} selection process of our framework.

\begin{table}[h]
\begin{center}
\begin{tabular}{|c|c|c|}\hline
Dataset & High recall words & Low recall words \\\hline
IAPR & court, tennis, cyclist, net,  & backpack, pavement, green, stand,\\
 &  cyclist, spectator &   hedge, garden, clothes\\
\hline
ESP & man, white, airplane, sky    & mouth, child, swim, statue, hill \\
 & coin,red, sky, black, & shadow, school,  floor\\
\hline
Flickr30K & wave, man, dog, soccer   & vest, striped, sunglass, arm, vest, \\
& soccer, shirt, dog & dark, watching, long\\
\hline
\end{tabular}
\end{center}
\vspace{-0.25cm}
\caption{Sample high and low recall words; In addition to high frequency words (`man', `sky' ), words related to visual {\it themes} ( {\em tennis match} or {\em bicycle race}) achieve better recall.}
\label{tb:recall}
\vspace{-0.25cm}
\end{table}

Number of clusters formed to represent the {\em theme} structure among a dataset seems to be an important hyper-parameter. We experimented with a wide range of numbers of clusters to see the effects of this hyper-parameter on the performance of the system. Our system employs hierarchical clustering with a cut-off threshold to form these clusters. This method leaves limited flexibility in the hands of the developer to change the numbers of generated clusters. Very small value of cut-off threshold results in the formation of single-member clusters corresponding to every data item. Very large value of cut-off threshold forces all data items to become members of a single cluster. Results shown in Tables \ref{tb:noofclusters} indicate that within the allowed range of numbers of clusters, the performance of our annotation system remains stable for all three datasets. Mean F-score varies very little for a wide change in the numbers of clusters.
\begin{table}[h]
\setlength{\tabcolsep}{4.0pt}
\begin{center}
\begin{tabular}{|c|c|c|c|c|}\hline
Dataset & No. of clusters & Mean Precision & Mean Recall & Mean F-score \\\hline
IAPR TC-12 & $1800$ & $41$	& $43$ & $42$\\\cline{2-5}
                  & $1250$ & $40$	& $42$ & $41$ \\\cline{2-5}
                  & $600$	& $43$	& $41$ & $42$\\\hline   
ESP            & $1800$ & $39$	& $38$ & $39$\\\cline{2-5}
                  & $1350$ & $39$	& $38$ & $39$\\\cline{2-5}
                  & $900$	& $39$	& $38$ & $39$\\\hline  
Flickr30K    & $4350$ & $25$	& $27$ & $26$\\\cline{2-5}
                  & $3300$ & $26$	& $27$ & $27$ \\\cline{2-5}
                  & $2900$ & $27$	& $26$ & $27$\\\hline  
\end{tabular}
\end{center}
\vspace{-0.25cm}
\caption{Effects of the numbers of {\em themes}/clusters on the performance of the system}
\label{tb:noofclusters}
\vspace{-0.25cm}
\end{table}

\subsubsection{Precision for descriptive words}

As explained in Section \ref{sc:intro}, %precision of a random classifier for a certain label is directly proportional to the frequency of that label. Since image annotation systems are basically classifiers with vocabulary words as their target label, 
most annotation systems have a tendency to be more precise for highly frequent words only. On the other hand, words with rare-to-moderate frequency are considered more important in search and retrieval scenario\cite{vsm_text_retrieval}. %Information theory explains this phenomenon as {\em ``surprising"} events, i.e., rare words, having more information content than {\em ``expected"} events defined by highly frequent words. Hence, most image annotation systems preform better for words of low information content than the words which are more important in search and retrieval scenario. 

\begin{figure}[h]
\begin{center}
\includegraphics[width = 0.65\textwidth]{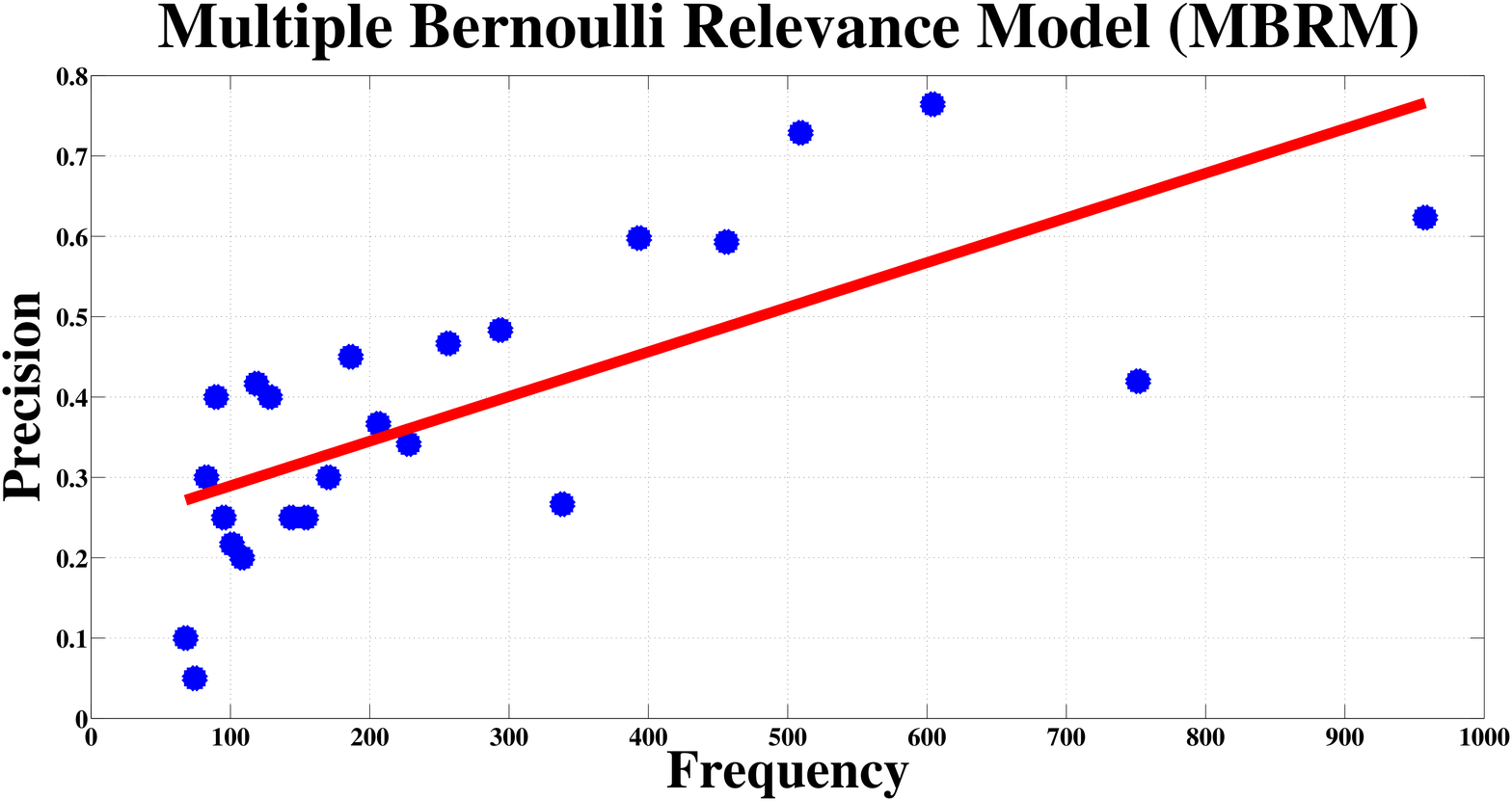}
\includegraphics[width = 0.75\textwidth]{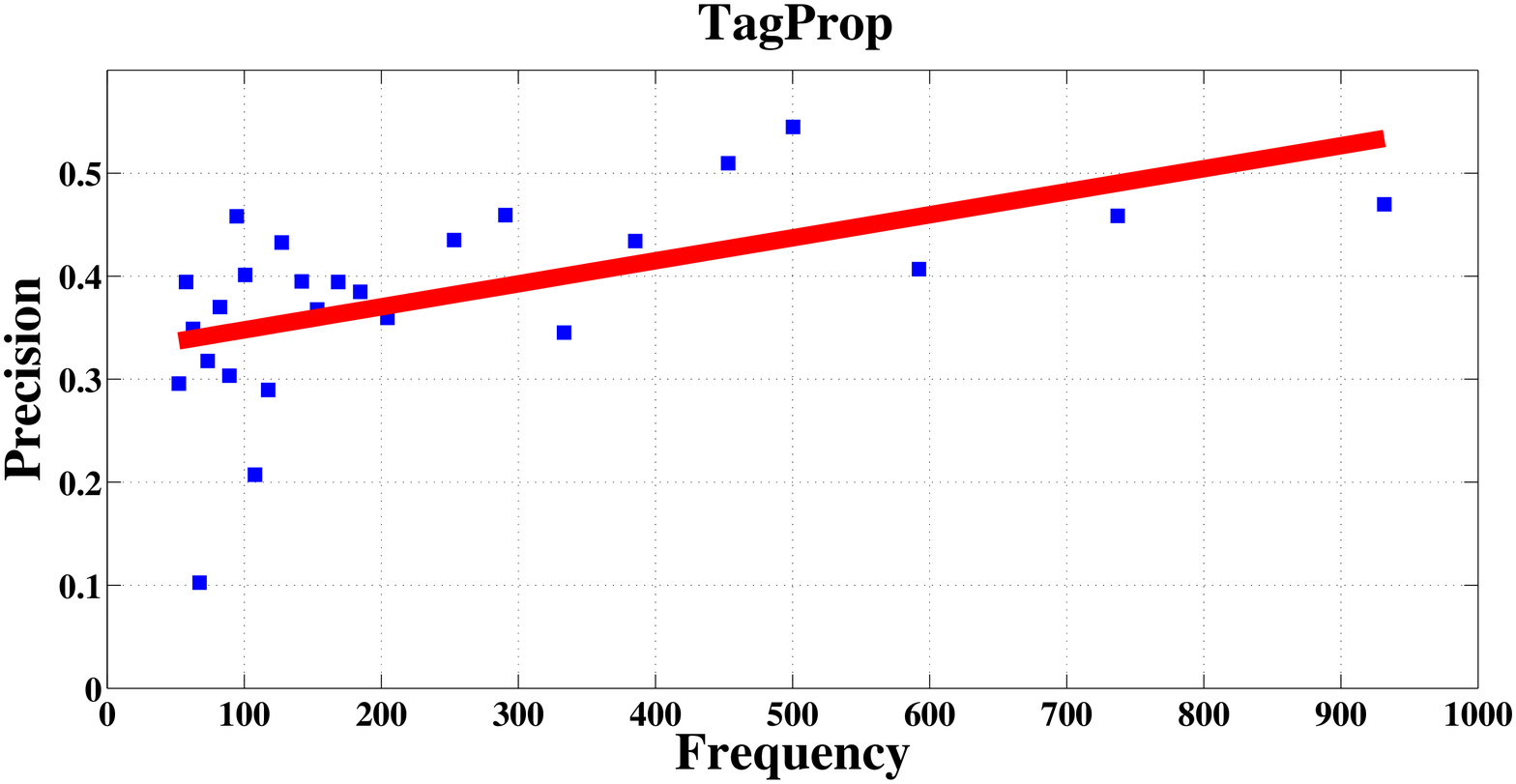}
\includegraphics[width = 0.65\textwidth]{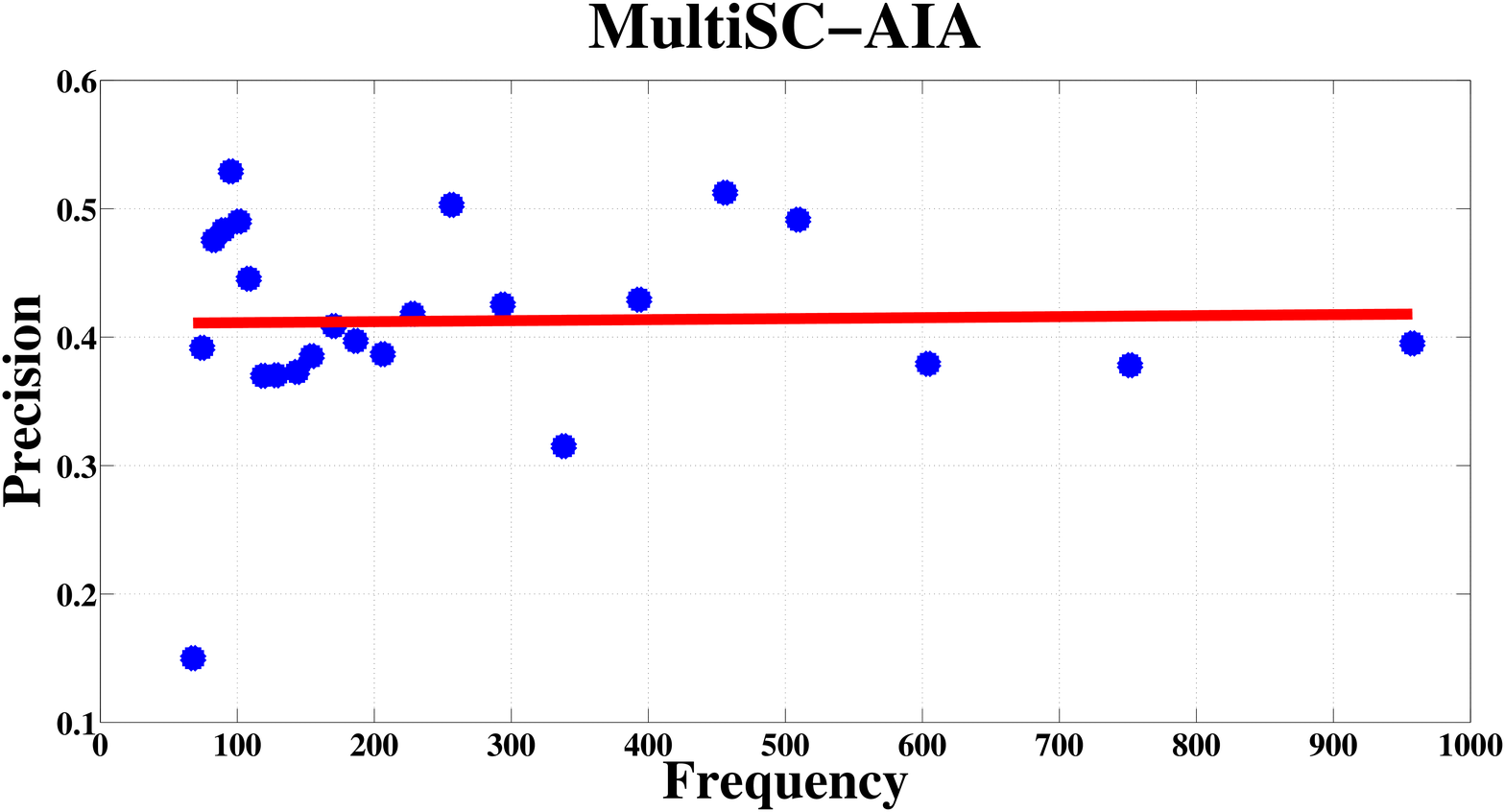}
\end{center}
\vspace{-0.5cm}
\caption{Effects of label frequency on the precision for different image annotation systems}
\label{fg:prec_freq}
\vspace{-0.25cm}
\end{figure}

Since the annotation system proposed in this paper explicitly attempts at achieving a balanced response in terms of precision and recall, it succeeds in maintaining its performance for words with a wide range of frequencies. We present precision-vs.-frequency plots for the proposed system as well as previously proposed MBRM and TagProp systems in Figure \ref{fg:prec_freq}, to highlight this favorable characteristic of the proposed system.

Annotations are sorted in the ascending order of frequency for IAPR TC-12 dataset. Mean precision and mean frequency of sets of every $10$ annotations are calculated. Lines fitted through the points representing these precision-frequency tuples (frequency along x-axis and precision along y-axis) are the precision-vs.-frequency plots. It is clear that our system maintains high precision for words with a wide range of frequencies. MBRM is highly precise for very frequent words only. The behavior of MBRM system is in compliance with the implications of Equation \ref{eq:rec2}. TagProp with sigmoidal modulation was designed to improve the performance of the system for rare words. It is evident that the difference between the precision of the system for very frequent and extremely rare words for TagProp is smaller than that of MBRM. Still, our system achieves the most stable performance over the whole word-frequency range.

The proposed system not only outperforms other annotation methods in overall mean precision and recall scores per word, but also performs much better than other systems for highly descriptive words of moderate frequency (`tennis', `stadium', `waterfall', `cathedral', `player'). Such words are more important for search and retrieval systems than highly frequent words (`sky', `man', `wall'). Most of the other systems tend to rely on precision of highly frequent words to improve overall performance.

%The proposed multi-layer sparse coding framework takes advantage of the visual representations learned by ImageNet-trained CNN which strongly correlates with ground truth annotation for various image annotation benchmark datasets. With the use of multiple layers of sparse coding the proposed framework overcomes an inherent problem in annotation models, i.e., low recall because of large number labels. Thorough experimentation clearly indicates that our system maintains its signature characteristics for a variety of annotation benchmark datasets.

\begin{table}[t]
\begin{center}
\begin{tabular}{|c|c|c|c|}
\hline Word  & Test    & Relevant training images  & Irrelevant training    \\
           ($v$)&Image  &       with word $v$           &  images of word $v$ \\
                  &            &                                      &    (Noise)                 \\\hline
Car & \includegraphics[width = 0.125\textwidth, height = 0.1\textwidth, trim=0 0 0 -25]{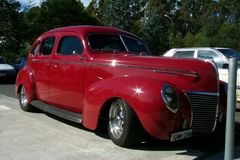} & 

\includegraphics[width = 0.125\textwidth, height = 0.1\textwidth, trim=0 0 0 -25]{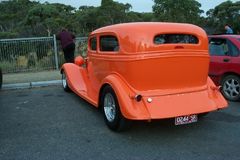} 
\includegraphics[width = 0.125\textwidth, height = 0.1\textwidth, trim=0 0 0 -25]{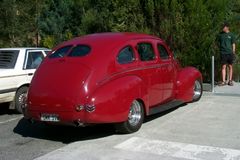}
\includegraphics[width = 0.125\textwidth, height = 0.1\textwidth, trim=0 0 0 -25]{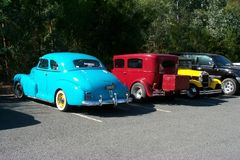} 
 & 
\includegraphics[width = 0.125\textwidth, height = 0.1\textwidth, trim=0 0 0 -25]{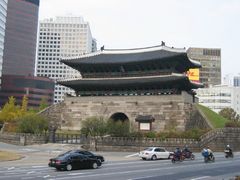} 
\includegraphics[width = 0.125\textwidth, height = 0.1\textwidth, trim=0 0 0 -25]{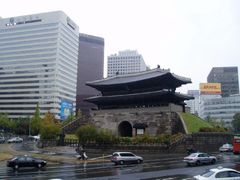}\\
% & & 
%\includegraphics[width = 0.15\textwidth, height = 0.1\textwidth, trim=0 0 0 -25]{Figures/Noise/car/good/4.jpg}
%\includegraphics[width = 0.15\textwidth, height = 0.1\textwidth, trim=0 0 0 -25]{Figures/Noise/car/good/5.jpg}
%&
%\includegraphics[width = 0.15\textwidth, height = 0.1\textwidth, trim=0 0 0 -25]{Figures/Noise/car/bad/3.jpg}
%\includegraphics[width = 0.15\textwidth, height = 0.1\textwidth, trim=0 0 0 -25]{Figures/Noise/car/bad/4.jpg} \\

\hline

Sea & \includegraphics[width = 0.125\textwidth, height = 0.1\textwidth, trim=0 0 0 -25]{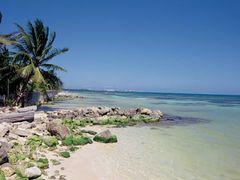} & 
\includegraphics[width = 0.125\textwidth, height = 0.1\textwidth, trim=0 0 0 -25]{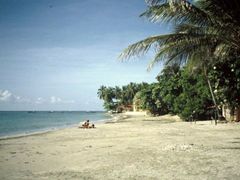} 
\includegraphics[width = 0.125\textwidth, height = 0.1\textwidth, trim=0 0 0 -25]{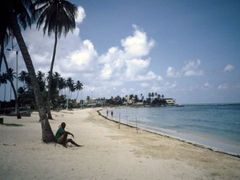}
\includegraphics[width = 0.125\textwidth, height = 0.1\textwidth, trim=0 0 0 -25]{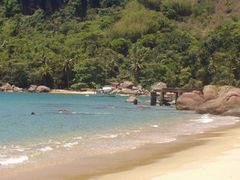}
&
\includegraphics[width = 0.125\textwidth, height = 0.1\textwidth, trim=0 0 0 -25]{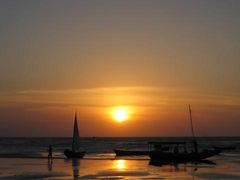} 
\includegraphics[width = 0.125\textwidth, height = 0.1\textwidth, trim=0 0 0 -25]{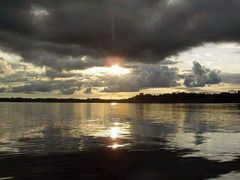}\\
%& & 
%\includegraphics[width = 0.15\textwidth, height = 0.1\textwidth, trim=0 0 0 -25]{Figures/Noise/sea/good/4.jpg}
%\includegraphics[width = 0.15\textwidth, height = 0.1\textwidth, trim=0 0 0 -25]{Figures/Noise/sea/good/5.jpg}
%&
%\includegraphics[width = 0.15\textwidth, height = 0.1\textwidth, trim=0 0 0 -25]{Figures/Noise/sea/bad/3.jpg}
%\includegraphics[width = 0.15\textwidth, height = 0.1\textwidth, trim=0 0 0 -25]{Figures/Noise/sea/bad/4.jpg} \\

\hline

%Building & \includegraphics[width = 0.125\textwidth, height = 0.1\textwidth, trim=0 0 0 -25]{Figures/Noise/building/building.jpg} & 
%\includegraphics[width = 0.125\textwidth, height = 0.1\textwidth, trim=0 0 0 -25]{Figures/Noise/building/good/1.jpg} 
%\includegraphics[width = 0.125\textwidth, height = 0.1\textwidth, trim=0 0 0 -25]{Figures/Noise/building/good/2.jpg}
%\includegraphics[width = 0.125\textwidth, height = 0.1\textwidth, trim=0 0 0 -25]{Figures/Noise/building/good/3.jpg}
%& 
%\includegraphics[width = 0.125\textwidth, height = 0.1\textwidth, trim=0 0 0 -25]{Figures/Noise/building/bad/1.jpg} 
%\includegraphics[width = 0.125\textwidth, height = 0.1\textwidth, trim=0 0 0 -25]{Figures/Noise/building/bad/2.jpg}\\
%%\includegraphics[width = 0.08\textwidth, height = 0.08\textwidth, trim=0 0 0 -25]{Figures/Noise/building/good/4.jpg}
%%&
%%
%%\includegraphics[width = 0.08\textwidth, height = 0.08\textwidth, trim=0 0 0 -25]{Figures/Noise/building/bad/3.jpg}
%%\includegraphics[width = 0.08\textwidth, height = 0.08\textwidth, trim=0 0 0 -25]{Figures/Noise/building/bad/4.jpg} \\
%\hline

Table & \includegraphics[width = 0.125\textwidth, height = 0.1\textwidth, trim=0 0 0 -25]{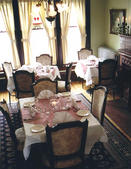} & 
\includegraphics[width = 0.125\textwidth, height = 0.1\textwidth, trim=0 0 0 -25]{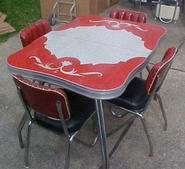} 
\includegraphics[width = 0.125\textwidth, height = 0.1\textwidth, trim=0 0 0 -25]{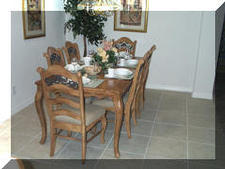}
\includegraphics[width = 0.125\textwidth, height = 0.1\textwidth, trim=0 0 0 -25]{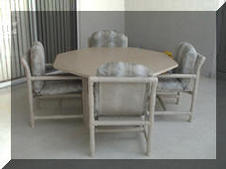}
&
\includegraphics[width = 0.125\textwidth, height = 0.1\textwidth, trim=0 0 0 -25]{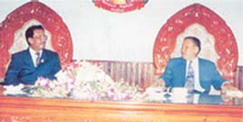} 
\includegraphics[width = 0.125\textwidth, height = 0.1\textwidth, trim=0 0 0 -25]{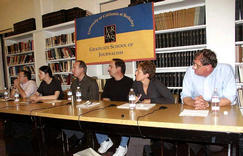}\\
%& &
%\includegraphics[width = 0.15\textwidth, height = 0.1\textwidth, trim=0 0 0 -25]{Figures/Noise/table/good/3.jpg}
%\includegraphics[width = 0.15\textwidth, height = 0.1\textwidth, trim=0 0 0 -25]{Figures/Noise/table/good/4.jpg}
%&
%\includegraphics[width = 0.15\textwidth, height = 0.1\textwidth, trim=0 0 0 -25]{Figures/Noise/table/bad/3.jpg}
%\includegraphics[width = 0.15\textwidth, height = 0.1\textwidth, trim=0 0 0 -25]{Figures/Noise/table/bad/4.jpg} \\
\hline

Airplane & \includegraphics[width = 0.125\textwidth, height = 0.1\textwidth, trim=0 0 0 -25]{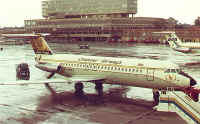} & 
\includegraphics[width = 0.125\textwidth, height = 0.1\textwidth, trim=0 0 0 -25]{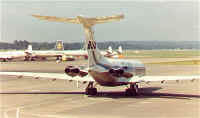} 
\includegraphics[width = 0.125\textwidth, height = 0.1\textwidth, trim=0 0 0 -25]{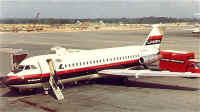}
\includegraphics[width = 0.125\textwidth, height = 0.1\textwidth, trim=0 0 0 -25]{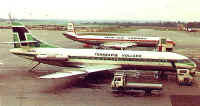}
&
\includegraphics[width = 0.125\textwidth, height = 0.1\textwidth, trim=0 0 0 -25]{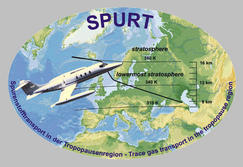} 
\includegraphics[width = 0.125\textwidth, height = 0.1\textwidth, trim=0 0 0 -25]{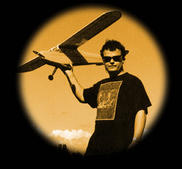}\\
%& &
%%\includegraphics[width = 0.08\textwidth, height = 0.07\textwidth, trim=0 0 0 -25]{Figures/Noise/airplane/good/3.jpg}
%\includegraphics[width = 0.15\textwidth, height = 0.1\textwidth, trim=0 0 0 -25]{Figures/Noise/airplane/good/4.jpg}
%\includegraphics[width = 0.15\textwidth, height = 0.1\textwidth, trim=0 0 0 -25]{Figures/Noise/airplane/good/5.jpg}
%&
%\includegraphics[width = 0.15\textwidth, height = 0.1\textwidth, trim=0 0 0 -25]{Figures/Noise/airplane/bad/3.jpg}
%\includegraphics[width = 0.15\textwidth, height = 0.1\textwidth, trim=0 0 0 -25]{Figures/Noise/airplane/bad/4.jpg} \\
\hline

\end{tabular}
\end{center}
\vspace{-0.25cm}
\caption{\label{tb:noise} Noise in training data for individual words (IAPR TC-12 dataset) }
\vspace{-0.25cm}
\end{table}

\subsubsection{Noise reduction}
Images of one {\it theme} share {\it distinctive} words in their descriptions, and have similar visual contents and CNN-learned representations. The first layer of sparse coding lets only images of relevant {\it themes} pass to the second layer of sparse coding, implying that the second layer processes only training images that are visually and semantically similar to the test image. 

In the second sparse coding layer of the proposed framework, a sparse coding model is learned for each word. One word does not strictly limit the variety of visual contents. The same word can be associated with vastly different images. Only of a subset of these images are relevant to the test image while others act as noise when linear regression model for the test image is being learned. In our framework, the first layer acts as a filter and removes images which are visually and semantically different from the test image, thus, making the next layer of sparse coding more effective by reducing the noise encountered by this layer. Table \ref{tb:noise} shows examples of images considered relevant to the given test image for a certain word, as well images deemed as noise under the same setting.

\subsubsection{Theme Selection and Image Organization}
\label{subsubsc:theme_selection}
%Group sparse modeling at the first layer of sparse coding module identifies $K^I$ {\it themes} as relevant to the test image $I$ such that $K^I > 1$, in general. Every {\it theme} whose member training images are assigned positive coefficients while minimizing the objective function described in equation \ref{eq:layer1} is considered relevant to the test image $I$.
As explained in Section \ref{subsubsc:sglasso}, a test image $I$ may be associated with multiple {\it themes}.  We observed that multiple {\it themes} associated with the given test image $I$ usually represent multiple aspects of that image. No one {\it theme} can describe the image in totality but the combination of these {\it themes} successfully pinpoint contents of the given image. 
\begin{figure}[h]
\hrule
\vspace{0.025in}
\begin{minipage}[t]{0.242\textwidth}
\centering
\includegraphics[width = 0.48\textwidth, height = 0.45\textwidth]{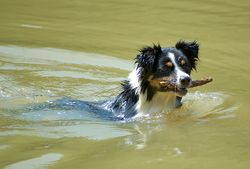}
\includegraphics[width = 0.48\textwidth, height = 0.45\textwidth]{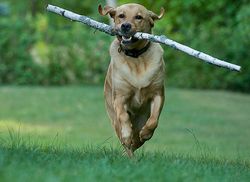}
\includegraphics[width = 0.48\textwidth, height = 0.45\textwidth]{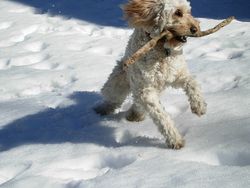}
\includegraphics[width = 0.48\textwidth, height = 0.45\textwidth]{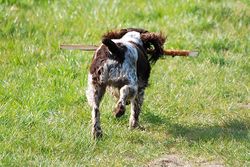}
\includegraphics[width = 0.48\textwidth, height = 0.45\textwidth]{Figures/theme_rel/theme1_mouth/5.jpg}
\includegraphics[width = 0.48\textwidth, height = 0.45\textwidth]{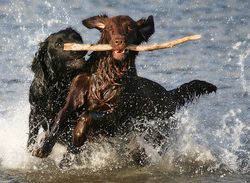}
{Theme$\#1$:}\\
{Dog with object}\\
{in his mouth}
\end{minipage}
\hfill \vrule \hfill
\begin{minipage}[t]{0.242\textwidth}\vrule
\centering
\includegraphics[width = 0.47\textwidth, height = 0.45\textwidth]{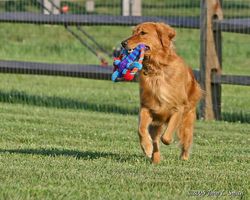}
\includegraphics[width = 0.47\textwidth, height = 0.45\textwidth]{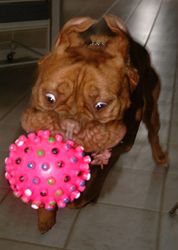}
\includegraphics[width = 0.47\textwidth, height = 0.45\textwidth]{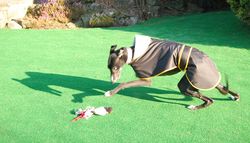}
\includegraphics[width = 0.47\textwidth, height = 0.45\textwidth]{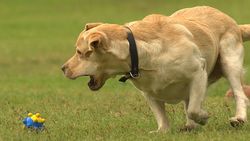}
\includegraphics[width = 0.47\textwidth, height = 0.45\textwidth]{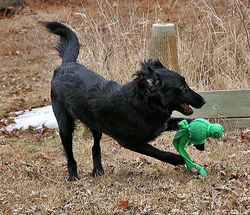}
\includegraphics[width = 0.47\textwidth, height = 0.45\textwidth]{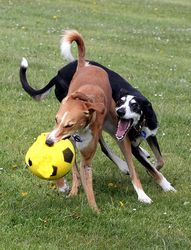}
{Theme$\#2$:}\\
{Dog with toy}
\end{minipage}
\hfill \vrule \hfill
\begin{minipage}[t]{0.242\textwidth}\vrule
\centering
\includegraphics[width = 0.47\textwidth, height = 0.45\textwidth]{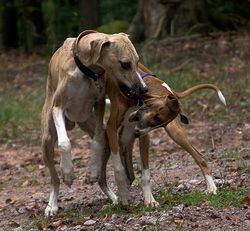}
\includegraphics[width = 0.47\textwidth, height = 0.45\textwidth]{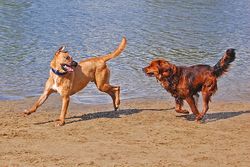}
\includegraphics[width = 0.47\textwidth, height = 0.45\textwidth]{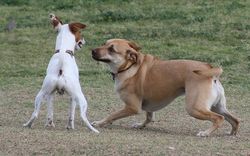}
\includegraphics[width = 0.47\textwidth, height = 0.45\textwidth]{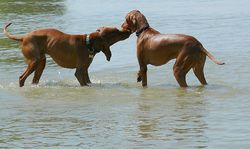}
\includegraphics[width = 0.47\textwidth, height = 0.45\textwidth]{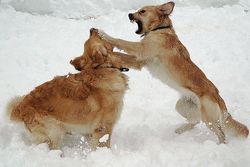}
\includegraphics[width = 0.47\textwidth, height = 0.45\textwidth]{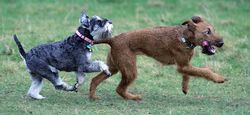}
{Theme$\#3$:}\\
{Two dogs}
\end{minipage}
\hfill \vrule \hfill
\begin{minipage}[t]{0.242\textwidth}\vrule
\centering
\includegraphics[width = 0.47\textwidth, height = 0.45\textwidth]{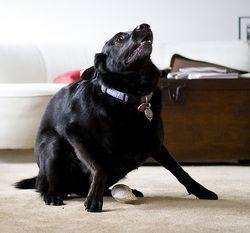}
\includegraphics[width = 0.47\textwidth, height = 0.45\textwidth]{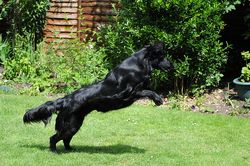}
\includegraphics[width = 0.47\textwidth, height = 0.45\textwidth]{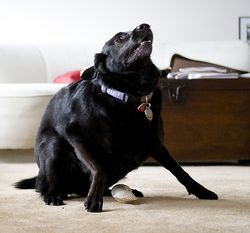}
\includegraphics[width = 0.47\textwidth, height = 0.45\textwidth]{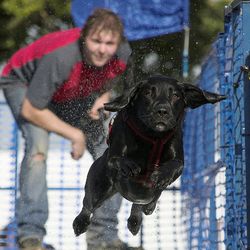}
\includegraphics[width = 0.47\textwidth, height = 0.45\textwidth]{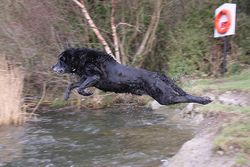}
\includegraphics[width = 0.47\textwidth, height = 0.45\textwidth]{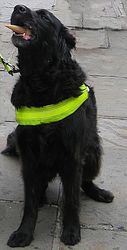}
{Theme$\#4$:}\\
{Black dog}
\end{minipage}
\vspace{0.025in}
\hrule
\vspace{0.025in}
%\begin{minipage}{0.13\textwidth}{Test image:}
%\end{minipage}
\begin{minipage}{0.14\textwidth}
\includegraphics[width = 1.0\textwidth]{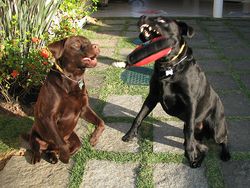}
\end{minipage}
\begin{minipage}{0.8\textwidth}
{The test image shows `two dogs with the black dog holding a }\\
{toy in his mouth'. Four related {\it themes} cover multiple aspects }\\
{of image contents, e.g., `two dogs', `black dog', etc. }
\end{minipage}
\vspace{0.025in}
\hrule
\vspace{0.025in}
\vspace{0.1cm}
\vspace{-0.25cm}
\caption{Sample images from multiple {\it themes} associated with the given test image}
\label{fg:theme_rel}
\end{figure} 

The selected {\it themes}, in addition to annotated words, have huge potential for image database organization and management systems. Images can be linked to each other through common {\it themes} for easy access. Each link will represent certain aspect of the visual contents. A new image may be integrated with the database in such a way that every aspect of visual contents is identifiable by multiple links based on multiple relevant {\it themes} identified through group sparse coding. Figure \ref{fg:theme_rel} shows an example of a test image and samples from four of its relevant {\it themes} where each {\em theme} identifies a certain aspect of contents of the test image.

\subsubsection{Time complexity}
Multilayer sparse coding framework reduces time complexity of the system as the second layer has to deal with subsets of training data and vocabulary sets. A single layer sparse coding framework is essentially a special degenerate case of our model obtained by assuming that all of the training data belong to one {\it theme}. Consequently, this one {\it theme} is appropriate for every test image $I$ and all of the training data is passed to the next layer of sparse coding ($\mathcal{J}^I = \mathcal{J}$ and $\mathcal{W}^I = \mathcal{W}$). The system then needs to learn a regularized linear regression model for every vocabulary word over all training images tagged with that word. Our experiments indicate that the processing time per image for such a case can be up to an order of magnitude more than the proposed multi-layer model, depending on regularization parameters. Annotation accuracy is also lower than our model for challenging datasets such as Flickr30K. Flickr30K dataset is larger and more diverse than IAPR TC-12 and ESP datasets, implying more noise in training set for each word.

\subsubsection{Analysis of Multi-Layer Sparse Coding Framework}

One prominent characteristic of the proposed system is that it maintains its precision performance for words or labels with a  wide range of frequencies while most of the previously proposed annotation systems tend to achieve better precision scores for high-frequency words or labels only. This property is a direct result of {\em theme}-based filtering of training data ($\mathcal{J}^I$) and vocabulary set ($\mathcal{W}^I$) at the first layer of sparse coding framework. %The first layer of group sparse coding identifies {\em themes} relevant to the test image and filters training and vocabulary sets to form modified training set $\mathcal{J}^I\subset\mathcal{J}$ and vocabulary set $\mathcal{W}^I\subset\mathcal{W}$. Sets $\mathcal{J}^I$ and $\mathcal{W}^I$ contain only training images and words belonging to {\em themes} relevant to the given test image $I$, respectively. 

As explained in Section \ref{subsubsc:theme}, {\em themes} are groups of training images sharing distinctive words with high {\em tfIdf} in their descriptions. Let $v_t$ denote a distinctive word  with high {\em tfIdf} (such as `cathedral' or `waterfall' for IAPR TC-12 dataset) while $v_c$ denotes a common word with high frequency (such as `man' or `sky'). $N_{v_t}$ and $N_{v_c}$ denote the numbers of training images tagged with words $v_t$ and $v_c$, respectively. Since $v_c$ is a high frequency word while $v_t$ is the distinctive property of a few images only, we can say that $N_{v_c} >> N_{v_t}$. Let us assume that word $v_t$ is among the set of distinctive words shared by the training images of {\em theme} $T$ and {\em theme} $T$ is relevant to the given test image $I$. In this case, all training images of {\em theme} $T$ are member of the filtered training set $\mathcal{J}^I$ and the word $v_t$ is member of filtered vocabulary set $\mathcal{W}^I$. On the other hand, many training images tagged with common word $v_c$ are filtered out of the training set $\mathcal{J}^I$. If $N^{I}_{v_t}$ and $N^{I}_{v_c}$ denote the numbers of images tagged with words $v_t$ and $v_c$ in the filtered set $\mathcal{J}^I$, respectively, then $N^{I}_{v_t} \approx N_{v_t}$ while $N^{I}_{v_c} << N_{v_c}$. 

The columns of the design matrix $A^{I_v}$ for test image $I$ given word $v$ are the training images tagged with word $v$, passed to the second layer of sparse coding. If design matrix $A$ is of size $n\times p$, $p=N^{I}_{v_c}$ for the common word $v_c$ and $p=N^{I}_{v_t}$ for {\em theme} word $v_t$, given the test image $I$.  We know that the quality of the model estimated by lasso is directly dependent on the number of predictor variables or $p$. As explained in \cite{highdimensionalstats}, 
\begin{equation}
\frac{||A^{I_v}(\hat{\mathbf{w}}-\mathbf{w}^{o})||_{2}^{2}}{n} \leq const.\frac{\sigma^{2}log(p)s_o}{n}
\end{equation}
where $\mathbf{w}^{o}$ denotes the true coefficient vector and $\hat{\mathbf{w}}$ denotes the estimated coefficient vector. $s_o$ is the size of the activation set or the number of predictors which are assigned non-zero coefficients.

The proposed framework maintains its performance for words with a wide range of frequencies as the values of $p$ for all vocabulary words are not as widely spread out for the second layer of sparse coding of the proposed framework as they are for the overall dataset. Many images tagged with highly frequent words are filtered out before the second layer of sparse coding, bring their corresponding $p$ value closer to that of words with moderate frequency. For previously proposed annotation systems, larger training data is available corresponding to highly frequent words. Hence, they are able to learn more precise models for highly frequent words than the highly descriptive words of moderate frequency. 

For the proposed framework, the design matrix $A^{I_v}$ may have high correlation among its columns as columns are training images that share {\em themes} in their visual and textual representation. Hebiri et al. thoroughly studied the effects of correlation in design matrix over lasso prediction\cite{lasso_correlation}. They concluded that lasso prediction works well for any degree of correlation with suitable tuning parameters. 

\section{Conclusion}
\label{sc:conclusion}

Automatic image annotation is a challenging yet important problem. A solution to this problem can greatly improve the performance of image search, retrieval, archival and organization systems. Most image annotation systems suffer from imbalance in precision and recall scores, and report high performance in terms of average precision per word by being precise for high frequency words only. Such words are less important in terms of search and retrieval than the words with moderate frequency. 

In this paper, we analyzed the properties of annotation systems as classifiers with vocabulary words as label. Our investigation concluded that simple random classifiers are prone to be highly precise for high-frequency words only while achieving low recall because of large number of labels available. Keeping these observations in mind, we propose a unique classifier that achieves a symmetric response with comparable precision and recall values while keeping both of them high. While our proposal is targeting the specific application area of image annotation \cite{Tariq_etal_2017_2,Tariq_etal_2017,tariq2013exploiting,tariq2015feature,tariq2014scene,tariq2015t}, the proposed principle of symmetric classifier has potential applications in diverse areas of image processing and computer vision, such as self-localization \cite{Junejo_etal_2010,junejo2008gps,Cao_Foroosh_2007}, surveillance \cite{Junejo_etal_2007,Junejo_Foroosh_2008,junejo2007trajectory}, action recognition \cite{Sun_etal_2015,Ashraf_etal_2014,Ashraf_etal_2013,Shen_Foroosh_2009,shen2008view,sun2011action,ashraf2014view,shen2008action,ashraf2010view,boyraz122014action,sun2014feature,ashraf2012human}, target localization and tracking \cite{Shu_etal_2016,shekarforoush2000multi,Milikan_etal_2017,millikan2015initialized}, shape description and object recognition \cite{Zhang_etal_2015,Cakmakci_etal_2008,Cakmakci_etal_2008_2,ali2016character}, image-based rendering \cite{alnasser2006image,balci2006real,balci2006image,shen2006video}, image restoration \cite{Foroosh_Chellappa_1999,Foroosh_etal_1996,shekarforoush19953d,lorette1997super,shekarforoush1998multi,berthod1994refining,shekarforoush1999conditioning,jain2008super,shekarforoush1999super}, and camera motion classification and quantification \cite{Junejo_etal_2011,Cao_Foroosh_2007,Cao_Foroosh_2006,cao2004camera,junejo2006calibrating,cao2006self,cao2006camera,ashraf2007near}, to name a few.

The proposed annotation system employs multiple layers of sparse coding treating training images as predictors and the test image as target signal. Training images are grouped in {\em themes} based on similarity in their distinctive textual components and visual similarity. The first layer takes advantage of this group structure among predictors and identifies {\em themes} relevant to the test image $I$. The next layer of sparse coding deals with reduced training and vocabulary sets, containing images and words belonging to the relevant {\em themes} only. This reduced vocabulary or labels' set results in balanced response in terms of precision and recall.  

While the idea of sparse coding has been used frequently for image processing problems of codebook generation and face recognition, our system employs a unique multilayer sparse coding framework and large application scope for a large labels' set problem, i.e., image annotation,  where each item is tagged with multiple labels. Our system performs systematic coarse-to-fine labeling without requiring coarse labels as prior knowledge. Coarse labels are automatically identified by our system in terms of {\em themes} present in available training data .

The proposed approach has been evaluated thoroughly over three popular benchmark datasets for automatic image annotation. The proposed system outperforms previously proposed annotation systems in terms of F-score which incorporates the trade-off between precision and recall. The system also maintains high precision for words with a wide range of frequencies. As explained in terms of information theory, less frequent words are more important in terms of search and retrieval than highly frequent words. Our system is highly precise for such less-frequent words as well. %In future, we intend to explore incorporation of additional sparse coding layers as well as comparative evaluation of different regularization terms.
%penalties included in their objective functions.




%--------------------------------------------------------------------------------------------- 

{%% \small
\bibliographystyle{plain}
\bibliography{refs,foroosh}
}

\end{document}